\documentclass[journal]{IEEEtran}

\usepackage{amsmath,amssymb,epsfig}
\usepackage{setspace}
\usepackage[final]{pdfpages}
\usepackage{cancel}
\usepackage[absolute]{textpos}
\usepackage{tikz}
\usetikzlibrary{shapes,arrows,petri,topaths}
\usepackage{tikz,fullpage}
\usepackage{tkz-berge}
\usepackage{caption}
\usepackage{subfig}
\usetikzlibrary{positioning}
\usepackage{enumerate}
\usepackage{mdframed}
\usepackage{gensymb}
\usepackage{float}
\usepackage{hyperref}
\usepackage[utf8]{inputenc}
\usepackage{nomencl}

\restylefloat{table}
\usepackage[linesnumbered]{algorithm2e}
\SetAlCapSkip{1em}
\usepackage[top = 54pt,left = 54pt,right = 54pt,left = 54pt]{geometry}

\newcommand{\QED}{$\blacksquare$}

\newcommand{\lb}{\left\lbrace}
\newcommand{\rb}{\right\rbrace}

\def\E{{\mathbb E}}

\IEEEoverridecommandlockouts                              % This command is only needed if 
\usepackage{graphicx} % for pdf, bitmapped graphics files

\makenomenclature

\newcommand{\beq}{\begin{equation}}
\newcommand{\eeq}{\end{equation}}
\newcommand{\beqa}{\begin{eqnarray}}
\newcommand{\eeqa}{\end{eqnarray}}
\newcommand{\beqan}{\begin{eqnarray*}}
\newcommand{\eeqan}{\end{eqnarray*}}

\newcommand{\R}{{\mathbb R}}

\newcommand{\BigO}[1]{\ensuremath{\operatorname{O}\left(#1\right)}}

\newcommand{\argmax}{\mbox{argmax}}

\newcounter{l1}
\newcounter{l2}
\newcounter{l3}
\newcommand{\bdotlist}{\begin{list}{$\bullet$}{}}
\newcommand{\bboxlist}{\begin{list}{$\Box$}{}}
\newcommand{\bbboxlist}{\begin{list}{\raisebox{.005in}{{\tiny
$\blacksquare$ \ \ }}}{}}
\newcommand{\bdashlist}{\begin{list}{$-$}{} }
\newcommand{\blist}{\begin{list}{}{} }
\newcommand{\barablist}{\begin{list}{\arabic{l1}}{\usecounter{l1}}}
\newcommand{\balphlist}{\begin{list}{(\alph{l2})}{\usecounter{l2}}}
\newcommand{\bAlphlist}{\begin{list}{\Alph{l2}.}{\usecounter{l2}}}
\newcommand{\bdiamlist}{\begin{list}{$\diamond$}{}}
\newcommand{\bromalist}{\begin{list}{(\roman{l3})}{\usecounter{l3}}}

\newcommand{\thm}[1]{\noindent \begin{theorem} #1   \end{theorem}}
\newcommand{\prop}[1]{\begin{proposition} #1 \end{proposition}}

\newcommand{\prf}[1]{{\em Proof:} \, #1 \hfill $\blacksquare$}

\newcommand{\defn}[1]{\begin{definition} {\rm #1 }
\end{definition}}

\newcommand{\cor}[1]{\begin{corollary}   #1  \end{corollary}}

\renewcommand{\theequation}{\arabic{equation}}
\newtheorem{theorem}{Theorem}[section]
\newtheorem{exercise}[theorem]{Exercise}
\newtheorem{lemma}[theorem]{Lemma}
\newtheorem{proposition}[theorem]{Proposition}
\newtheorem{corollary}[theorem]{Corollary}
\newtheorem{definition}[theorem]{Definition}
\newtheorem{remark}[theorem]{Remark}
\newtheorem{example}[theorem]{Example}

\title{\LARGE \bf
An Extended Treatment of Uncertainty Constrained Robotic Exploration: An Integrated Exploration Planner
}

\author{Alexander Ivanov$^{1}$ and Mark Campbell$^{2}$% <-this % stops a space
\thanks{$^{1}$information:{\tt\small  \hspace{5pt} A. Ivanov and M. Campbell are with the Sibley School of Mechanical and Aerospace Engineering, Cornell University, Ithaca, NY, 14850 USA. email: aii4@cornell.edu,mc288@cornell.edu }}%
\thanks{$^{2}$ Corresponding author}

\thanks{Funding:{\tt\small  \hspace{5pt} AFOSR Grant FA9550-12-1-0410:Dr. Fariba Farhoo program manager. }}%
}

\begin{document}

\maketitle
\thispagestyle{empty}
\pagestyle{empty}

%%%%%%%%%%%%%%%%%%%%%%%%%%%%%%%%%%%%%%%%%%%%%%%%%%%%%%%%%%%%%%%%%%%%%%%%%%%%%%%%
\begin{abstract}
Efficient robotic exploration of unknown, sensor limited, global-information-deficient environments poses unique challenges to path planning algorithms. In these difficult environments, no deterministic guarantees on path completion and mission success can be made in general. Integrated Exploration (IE), which strives to combine localization and exploration, must be solved in order to create an autonomous robotic system capable of long term operation in new and challenging environments. This paper formulates a probabilistic framework which allows the creation of exploration algorithms providing probabilistic guarantees of success. A novel connection is made between the Hamiltonian Path Problem and exploration. The Guaranteed Probabilistic Information Explorer (G-PIE) is developed for the IE problem, providing a probabilistic guarantee on path completion, and asymptotic optimality of exploration. A receding horizon formulation, dubbed RH-PIE, is presented which addresses the exponential complexity present in G-PIE. Finally, RH-PIE planner is verified via autonomous, hardware-in-the-loop experiments. 
\end{abstract}

\nomenclature{$x_k, \bar{x}_k, \hat{x}_k$}{True, mean, and reference robot state at $k$}
\nomenclature{$B_{\mathrm{pos}}$}{Robot state error cost for RH-PIE}
\nomenclature{$B_{\mathrm{info}}$}{ Information reward for RH-PIE}
\nomenclature{$X_k$}{Robot state random variable}
\nomenclature{$X_{t:T} \in \mathcal{X}_{t:T}$}{A path from $t$ to $T$, and its allowable set}
\nomenclature{$Z_{t:T}$}{Concatinated measuremtns from time $t$ to $T$}
\nomenclature{$R(\centerdot)$}{Reward function mapping a path to $\mathbb{R}$}
\nomenclature{$\E[\centerdot]$}{Expectation}
\nomenclature{$H(\centerdot)$}{Entropy}
\nomenclature{$p(\centerdot), P(\centerdot)$}{Probability density and mass}
\nomenclature{$\mathcal{C}$}{Cardinality}
\nomenclature{$C_i$}{The $i$th 2D grid cell}
\nomenclature{$C_{\mathrm{free}}, C_{\mathrm{obs}}$}{Free and obstacle configuration spaces}
\nomenclature{$tr(\centerdot), \lambda_1(\centerdot)$}{Trace and spectral norm of a matrix}
\nomenclature{$\mathcal{G}(V,E)$}{A graph with vertex and edge sets}
\nomenclature{$M(t)$}{Set of RVs representing a map at time $t$}
\nomenclature{$M(\centerdot)$}{Subset of map RVs affected as a funciton of the argment}
\nomenclature{$\gamma \in \R^{+}$}{A constraint on the covariance norm }
\nomenclature{$\delta, \alpha \in (0,1)$}{Constraint parameters defining proability of localization }
\nomenclature{$n_x, n_z, n_m$}{Dimentions of robot state, measurement, and map}
\nomenclature{$T, T_1$}{Time horizon variables}
\nomenclature{$\beta \in (0,1)$}{Scalerization parameter}
\nomenclature{$\Sigma_k \in \Re^{n_x \times n_x}$}{Robot covariance at time $k$}
\nomenclature{$v_i, e_i$}{Indexed vertex and indexed edge}
\nomenclature{$\mathcal{L} \subset \R^2$}{Localization region}
\nomenclature{$\phi,\theta \in (0,1)$}{Mis-detection and false alarm parameters}
\nomenclature{$\mathcal{N}(x, \Sigma)$}{Normal distribution with mean $x$ and covarience $\Sigma$ }
\nomenclature{$i$}{Index variable accosiated with position}
\nomenclature{$k$}{Index variable associated with descrete time}

\printnomenclature[55pt]

%\doublespacing
%%%%%%%%%%%%%%%%%%%%%%%%%%%%%%%%%%%%%%%%%%%%%%%%%%%%%%%%%%%%%%%%%%%%%%%%%%%%%%%%
\section{INTRODUCTION}
Exploration in sensor limited, global-information-deficient environments is necessary for mobile robots to act independently. Applications with these challenges include persistent robots new to a home or office, or robots operating in dangerous places where human participation is impossible and GPS coverage is limited. Examples of such scenarios include burning buildings and nuclear power plants. Any robot operating in these environments needs to have knowledge of its surroundings to fulfill higher level tasks.

%For example, a survey robot may need to explore a collapsed building to help operators asses the stability of the surrounds, or a reconnaissance robot in an urban canyon may need to search for targets of interest. 

The problem of exploring an unknown area while trying to complete a higher level mission is termed Integrated Exploration (IE) \cite{Makarenko}. Mission success in IE is formally defined as the ability of a robot to gather sensory information about its surroundings, while maintaining pose and/or maintaining the ability to recover from pose degradation. These two conflicting goals, along with the probabilistic nature of sensing, make IE a challenging problem.

\begin{figure}

\centering
t\includegraphics[scale=.4]{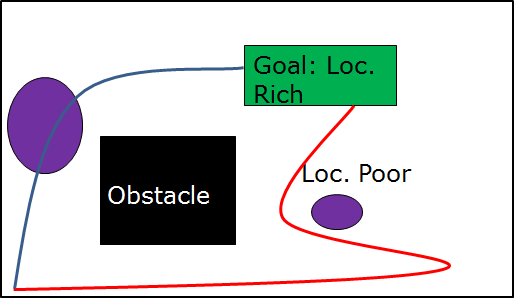}
\caption{Conceptual exemplification of the partially known IE problem: the robot must navigate from the start point to the goal region, avoid obstacles, and maintain a globally consistent pose estimate. It can gather some localization information in the purple regions, and rich localization information in the goal region.}
\vspace{-1.5em}
\label{fig:examplePath}
\end{figure}

Fig. (\ref{fig:examplePath}) shows a conceptual example of integrated exploration in a partially known environment. The robot initializes at the bottom left, with a goal of exploring its surroundings. In addition, the robot must maintain a consistent pose estimate along the way (i.e. not get lost). Since pose information is sparse, the robot must eventually travel to the green goal region, who's rich pose information enables the pose uncertainty to shrink to desired levels. In practice, this goal could be an area with well known landmarks, a distinct doorway within a building, or WiFi/GPS coverage. It is assumed that the robot knows the location of regions which are impassible (black), contain no pose information (white), and poor pose information (purple). In the example shown in Fig. (\ref{fig:examplePath}), the robot chooses between, two possible paths. Although the red path is longer and explores more area, the expected pose information is sparse. Conversely, the blue path explores less, but has much more pose information. The key research question addressed in this paper is: How can the robot optimize the exploration objective with a constraint of maintaining a globally consistent pose.

Makarenko et al. study a planner with information gain as a path reward \cite{Makarenko}. Paths are generated towards a `Frontier'  which differentiates between visited and unvisited areas of a map. A similar approach is taken by Freda et. al. \cite{Freda} where a set of randomized points is evaluated at each time step for their information and localization potential. These points serve as greedy next positions. Stachniss and Burgard explore a combination of a heuristic `closest location,' and a location with ``maximum information gain" to determine a one-step look-ahead path \cite{Stachniss,Stachniss2}. In each of these works, information-greedy paths are generated which do not consider the long term reward of their decisions. In addition, they provide no guarantee of successful completion of either the exploration or localization goals.

Martinez-Cantin uses the Minimum Mean-Square Error (MMSE) of a robot's pose  and a set of localizing features as a path cost \cite{Martinez}. A set of controllers is then used to execute this MMSE path. Sim and Roy consider an exploration algorithm to optimally reduce the uncertainty of landmarks through a breadth first search of a gridded map \cite{Roy2}. Navigation function techniques incorporate considerations such as terminal path uncertainty,  distance from obstacles, and localization \cite{Roy, Tovar,Julia}. Although these solutions consider path execution, re-localization, and longer look-ahead distance than do greedy methods, they provide no guarantees on completion of paths or their exploration objective.

Lunders, Levine, and How build on the Rapidly Exploring Random Tree (RRT) algorithm to generate a chance constrained subset of paths from a start position to a goal. These paths ensure a high probability of obstacle avoidance \cite{ChanceRRT}. Although collision avoidance is assured, mission goals associated with information gathering tasks are either not considered or not guaranteed \cite{Levine}. Carlone and Lyons create an MPC planner in an environment with known obstacles and bounded disturbances, and provide guarantees on collision avoidance while attempting to completely explore a space in an frontier sense \cite{Carlone_Kullback, Carlone_Integer}. Here, the speed of information gathering is not considered directly, which implies that the robot may take a prohibitively long time to explore the area. 

In summary, current work attempts to solve IE using either unconstrained reward functions or chance constrained methods for obstacle avoidance. Unconstrained methods enable information gathering, but make no guarantees. Chance constrained methods guarantee path completion but tend to explore a space slowly \cite{ChanceRRT, Levine}.

This work builds on the Guaranteed Probabilistic Information Explorer (G-PIE) in \cite{ivanov2016efficient}. The G-PIE is an information theoretic path planning algorithm that provides a solution to the IE problem with the characteristics of fast exploration \textit{and} guarantees on re-localization.

This work first defines the Integrated Exploration (IE) problem, along with the specific sub-problem of information gathering in partially known environments, the focus of this paper, in Sections \ref{sect:Framework} and \ref{sect:PartiallyKnown}. An information reward function is introduced followed by two important novelties: a bound is derived on the proposed information reward function, and a probabilistic re-localization constraint is developed in Section \ref{sect:Bound} and \ref{sect:LocConst}. These components enable the key attributes of the G-PIE formulation: 1) fast exploration; 2) asymptotically optimal exploration; 3)  a probabilistic guarantee of localization. Finally, expanding on \cite{ivanov2016efficient}, a receding time horizon implementation of the G-PIE is developed which allows near-real time planning in Section \ref{sect:RHPIE}. The behavior of the receding horizon G-PIE is analyzed via fully autonomous hardware-in-the-loop experiments in Section \ref{sect:Results}.

\section{General Framework for Integrated Exploration with Chance Constraints} \label{sect:Framework}

A framework for robotic path planning with information collection goals and probabilistic constraints is formulated with two major components: a reward function which facilitates information collection, and chance constraints which provide probabilistic guarantees on a problem-dependent goal. 

In a 2D environment, the robot pose at time $t$, $X_t$, consists of the 2D coordinates of a robot and its orientation. A path starting at time $t$ and terminating at time $T$ is denoted $X_{t:T}$ and the optimal path, $X^*_{t:T}$, maximizes a reward, $R(X_{t:T})$:

\begin{equation} \label{eq:Optim}
X^*_{t:T} = \argmax_{X_{t:T}} R(X_{t:T}).
\end{equation}
\noindent
Note that, $X_{t:T}$, is a random variable and has a distribution. In this work the terms path and path distribution are used interchangeably.% when there is no chance for confusion. 

Information collection tasks, such as exploration, can often be encoded as a function of Random Variables (RVs). Thus, quickly gathering information about these variables constitutes fast exploration. Several well known metrics which capture uncertainty of RVs are Fisher Information (FI), Kullback-Leibler Divergence (KLD), and  Differential Entropy (DE) \cite{Bishop06}. When considering an $n_m$ dimensional joint distribution, FI can be computationally expensive ($O(n_m^3)$) and require large amounts of storage ($O(n_m^2)$) \cite{Bishop06}. KLD measures the difference between two distributions. This is not sufficient to ensure information gain, however, because, although two distributions might differ drastically before and after an observation, their overall uncertainty may be similar (e.g covariance). In contrast to KLD, entropy provides an intuitive information-theoretic metric. A random variable has high entropy when it is very uncertain, and the entropy of a RV monotonically decreases as the RV becomes more certain. For these reasons, entropy is used as the information metric in this work. 

RVs are denoted by capital variables and their realizations are lowercase. The domain of a random variable such as $X_t$, is denoted by capital script $\mathcal{X}_t$. Therefore, the DE of a RV, such as $X_t$, is defined as:
\begin{equation}
H(X_t) =  - \int\limits_{x \in \mathcal{X}_t}  p(x) \log(p(x)) dx.
\end{equation} 
 In IE, a vector valued RV can represent quantities of interest, such as uncertain target positions or obstacles. As one becomes more certain of the true value of a RV, the corresponding DE $ H(\centerdot) \to - \infty$ in the continuous case and 0 if the RV is discrete. For the IE problem, most variables of interest are contained in a vectorized map variable $M(t)$ which has been sensed by measurements $Z_{0:t}$. An information theoretic reward function can be defined using entropy as:
\begin{equation}\label{eq:C}
R(X_{t:T}) = H(M(t)) - \underset{Z_{t:T}}{\E} \left[ M(T)) \right].
\end{equation}
The above equation is termed mutual information. Here, $Z_{t:T} = \lb z_{t}, ...,z_T \rb$ is the set of measurements taken of $M$ from time $t$ until some terminal time $T$ after which no further measurements are taken (i.e. path completion time). The dependence of $M(t)$ on $Z_{0:t}$ is suppressed in Eq. (\ref{eq:C}) to reduce clutter and should be read as $(M(t)|Z_{0:t})$. Eq. (\ref{eq:C}) rewards information gain. 

Chance constraints are a form of soft constraint which guarantees a given condition with high probability. For IE, chance constraints are assumed to be represented by a vector function $f(\centerdot)$  and a constant vector $w$. These constraints take the form:

 \begin{equation} \label{eq:GenConstraint}
  f(X_{t:T}) \leq w.
 \end{equation}
 
In the case of exploration, the distributions on the random variables associated with the robot location (i.e. the path) encode many variables such as variability in $M(t)$, availability of location information, sensor accuracy, etc. Examples of practical constraints include obstacles to avoid and uncertainty in robot location.  The general framework for information theoretic goals with guarantees via chance constraints is given as Eq. (\ref{eq:C}) \& (\ref{eq:GenConstraint}).

\section{EXPLORATION IN PARTIALLY KNOWN ENVIRONMENTS} \label{sect:PartiallyKnown}

This section provides a formulation of the Integrated Exploration problem which enables probabilistic guarantees on relocalization when obstacles in the environment are already known.

\subsection{List of Assumptions}

\begin{itemize}
	\item[1] At least one region, $\mathcal{L}$ ,with `good' pose information is assumed known. There are many scenarios in which such a region is known \textit{a priori} such as when distinct doorways in a building are known from blueprints. 
	\item[2] The obstacle and free spaces $C_{\mathrm{obs}}$ and $C_{\mathrm{free}}$ are known. This assumption is valid when an exploration area has been previously mapped but richer information, such as the location of structural damage around buildings, is required.
    \item[3] The space is represented as a 2D grid composed of  $n_m$ probabilistically independent grid cells $C_i$, a standard assumption in exploration problems \cite{thrun2005probabilistic}.
	\item[4] The allowable set of reference paths are node orderings from a mathematical graph $\mathcal{G}(V,E)$, and reference paths can be well followed by a low-level controller. This assumption has been used to great effect in recent literature such as \cite{Prentice}. 
	\item[5] The reference paths are required to be \textit{simple}, i.e. nodes can only be visited once within a path.  For information exploration problems, expected entropy is subject to diminishing returns with more measurements (Fig. \ref{fig:symetric}). Therefore, restricting paths to be simple is a reasonable assumption.
\end{itemize}

\subsection{Problem Formulation}
The primary focus of this paper is the IE problem in \textit{partially known} environments. Colloquially, the robot can be seen as a `tourist' who has a high level overview of an area, but still wants to explore this area in detail. This scenario implies knowledge of the location of impassible obstacles and localization areas and their accuracy. Localization Poor Areas (LPAs) are regions with highly uncertain localization information. Examples of LPAs include prior mapped SLAM landmarks or areas with known unique features which are difficult for the robot to detect. For example, computer vision feature detection algorithms are error prone. Location Rich Areas (LRAs) are regions with highly certain location information such as well known mapped landmarks,  WiFi regions, or areas in which GPS is available. Formal definitions of these regions are given shortly. 

The robot is tasked with gathering information about the location of `areas of interest' without getting lost along the way.  In a realistic scenario, the robot may be searching for wounded soldiers on the ground or potential victims at the windows of damaged buildings. The robot plans paths from its current location through the environment to search for areas of interest. The robot is constrained to terminate any planned paths inside the LRA, $\mathcal{L}$, with high probability. This requirement ensures the robot's ability to localize and continued exploration. 

Given the assumptions, it follows that:

\begin{equation} \label{eq:CSPLAM}
\underset{Z_{t:T}}{\E} \left[ H(M(T))\right] = \underset{Z_{t:T}}{\E} \left[ -\sum\limits^{n_m}_{i=1} \sum_{j=0}^1 P(C_i = j) \ln(P(C_i = j))\right]
\end{equation}

Here $M(t)$ encodes the location of areas of interest. The measurements $Z_{t:T}$ implicitly depend on the path realization $x_{t:T}$, but dependence is suppressed to avoid notational clutter. An `interesting' cell takes value $c_i =1$, while an uninteresting cell takes value 0. 

Several computational challenges arise, which are addressed by assumption 4. The distribution of the robot's path, $X_{t:T}$, is assumed to follow a deterministic reference trajectory $\hat{x}_{t:T}$ closely enough to allow a certainty equivalence approximation, $x_{t:T} \approx \hat{x}_{t:T}$ \cite{Prentice}. The realization of $\hat{x}_{t:T}$ is presented shortly. Without this assumption, Eq. (\ref{eq:CSPLAM}) requires integration over all possible robot states $\{x:p(X_t =x)>0 \}$, which is not tractable in general.

\subsection{Localization Constraint: Probabilistic Guarantees} \label{sect:LocConst}
The proposed information explorer must search for areas of interest over an extended time period; this implies the robot must robustly execute many paths over time. Given that the exploration region has sparse localization information, the robot must adequately recover pose as it moves about the space. A constraint is proposed to ensure that the robot is probabilistically guaranteed to recover adequate pose. A metric is first defined to determine the level of pose information in an area, which in turn enables the definition of regions of rich and poor localization information (LRA, LPA). A location is considered a rich source of localization information if a robot can reduce its pose uncertainty to an acceptable level by taking pose measurements. Formally,

\defn{\label{def:localizable} A point $x$ is $\gamma$ \textit{Localizable} \textbf{at time instance} $k$ if $$tr(\Sigma_k) < \gamma, \hspace{10pt} X_k = x $$}

\noindent
 By setting $\gamma$ small, the user requires a more accurate pose estimate. 

Consider now a path $X_{t:T}$ with a terminal location $X_T$. If the robot's pose uncertainty is large near the terminal point of the path, $X_T$, the robot must then localize before planning a new path. This desired behavior is described as a terminal constraint:

\defn{\label{def:GammaDelta} A realized point $x$ satisfies a $(\gamma, \delta)$ \textit{Localization Constraint} if $$P \left( \min_{t \in [T,\infty]} [tr(\Sigma_t)] < \gamma | X_T = x \right) \geq 1-\delta$$}.
\vspace{.5em}

\noindent
 By setting $\delta$ small, the user specifies a tighter constraint on the certainty of satisfying the localization metric. 

The set of all $(\gamma, \delta)$ feasible points is defined as $\mathcal{L}$, or the LRA. This set can be seen as a `re-localization' region where the robot can reduce its pose uncertainty to acceptable levels. This constraint implies that if a robot remains at position $x$, it has a $1-\delta$ chance of recovering adequate pose certainty, defined by $\gamma$. Given that the robot completes, and replans, many such exploratory paths during a mission, these LRAs and constraints give a formal confidence on the ability of the robot to explore over an extended period of time. In this work, $\mathcal{L}$ is assumed to be known; $\mathcal{L}$ can be calculated numerically by discretization as shown in \cite{Makarenko}. Furthermore, it is assumed that $\mathcal{L}$ can be represented by a set of convex polygons for ease of computation. This is a mild assumption since all polygonal regions can be triangulated and sets of polygons can approximate most practical regions.
	
The localization constraint in Def. (\ref{def:GammaDelta}) is challenging to compute; therefore, a stricter sense of localization is considered which proves to be more manageable:

\defn{ \label{def:AlmostSureLoc}  A point $x$ is \textit{Almost Surely} $\gamma$ \textit{Localizatiable} if $$\lim_{t \to \infty} P \left(tr(\Sigma_t) < \gamma | X = x \right) = 1$$}.
\vspace{.5em}

\noindent
The limit in Def. (\ref{def:AlmostSureLoc}) has been found to have a closed form solution in the EKF case \cite{Makarenko}. Given these notions of `localizability', a practical and tractable \textit{path} constraint which probabilistically guarantees localization is defined as: 
\defn{\label{def:LocFeas} A path, $X_{t:T}$, is \textit{Localizably Feasible} (LF) if $$P(X_T \in \mathcal{L}) \geq \alpha$$ }
\vspace{-1.5em}

\noindent
This constraint assures that a robot can adequately, with user defined probability $\alpha$, localize in a $(\gamma, \delta)$ sense upon path completion. The general problem is now given as:

\begin{equation} \label{eq:GnereralOptimization}
\begin{aligned}
& \underset{X_{t:T} \in \mathcal{X}_{t:T}, \hspace{3pt} T}{\text{maximize}}
& & R(X_{t:T}) \\
& \text{subject to}
& & P(X_T \in \mathcal{L}) \geq \alpha.
\end{aligned}
\end{equation}

\subsection{Path Representations and Optimization}

Given the formulation in Eq. \eqref{eq:GnereralOptimization}, two components are required to complete the formulation: 1) the form of $\mathcal{X}_{t:T}$ ; 2) the optimization strategy. 
\subsubsection{Path Generation} \label{PathGen}
Discrete sampling methods are used to generate reference paths for several reasons. First, sampling methods allow a detailed set of paths to be created while using a relatively small number of sample points. Second, sampling methods have \textit{complete} variants; i.e. variants which guarantee approximating any path between two points infinitely well as the number of sample points grows \cite{Frazzoli}.

A path graph, denoted $\mathcal{G}(V,E)$, is defined by a set of sampled vertices $V$ in $C_{\mathrm{free}}$, and edges, $E$, connecting elements of $V$. A reference trajectory generated from the graph $\mathcal{G}$ is piece-wise-linear, and denoted $\hat{x}_{t:T} = \lb \hat{x}_{k_1}, ..., \hat{x}_{k_2} \rb$, where each $\hat{x}_k \in V$ and $(\hat{x}_k,\hat{x}_{k+1}) \in E$. The trajectory of the robot $X_{t:T} \in \mathcal{X}_{t:T}$ is therefore a function of the reference trajectory $\hat{x}_{k_1:k_2}$. A variant of the Probabilistic Road Map (PRM) algorithm is used to generate the graph, $\mathcal{G}$, which represents the allowable class of piece-wise-linear reference paths. The PRM algorithm is desirable because it allows for fast computation of the proposed pose constraint given in Def. (\ref{def:LocFeas}) \cite{Prentice}. 

\subsubsection{LF Path Calculation} \label{sect:PathConstraintCalc}

Each potential path, $X_{t:T}$, must be evaluated as to whether it is an LF path as defined in Def. (\ref{def:LocFeas}). This is accomplished via predictive simulation. The simulated robotic system attempts to follow the reference trajectory $\hat{x}_{k_1:k_2}$ using a trajectory following controller and an EKF filter. The generated controls are used to forward-propagate the simulated robot's distribution, while expected `average' measurements from LPAs along the reference trajectory are used to update the simulated EKF covariance matrix \cite{thrun2005probabilistic}. Finally, the terminal distribution of $X_T$ is evaluated in reference to $\mathcal{L}$ to obtain a probability of relocalization. For further details on the simulation process, see \cite{Prentice}. Note that the terminal covariance associated with each path can be efficiently calculated by employing the `one-step' covariance update matrices described in \cite{Prentice}. The terminal localization constraint is then written as:

\begin{multline}
\label{eq:TerminalConstr}
f(X_{t:T}) = P(X_T \in \mathcal{L}) = \\ \int\limits_{x_T \in \mathcal{L}}   {\exp\left( (x_T-\bar{x}_T)^T \Sigma_T^{-1}(x_T-\bar{x}_T)  \right) \over \sqrt{(2 \pi )^{n_x} |\Sigma_T|}}dx_T.
\end{multline}

\noindent
where $\bar{x}_T$ and $\Sigma_T$ are the mean and covariance of the robot pose at the end of path simulation, and $n_x$ is the dimensionality of $X$. This integration may be difficult to compute for an unstructured LRA $\mathcal{L}$. However, restricting $\mathcal{L}$ to be of non-zero measure, Eq. (\ref{eq:TerminalConstr}) can be quickly evaluated by sampling $\mathcal{N}(\bar{x}_T, \Sigma_T)$, and checking for inclusion in $\mathcal{L}$. The probability in Eq. \eqref{eq:TerminalConstr} is then approximated as the proportion of samples found inside $\mathcal{L}$.  This probability can be computed to any precision desired at a cost of of the number of samples: $\mathcal{O}(n)$. 

\subsubsection{Optimality of Exploration Algorithms}
The goal of the optimization in this problem is to maximize a reward not minimize a cost. Because the optimization is performed over $\mathcal{G}$, this is a longest path problem \cite{Kaerger}. 

\thm{ \label{thm:appxLongest} If a polynomial time approximation algorithm exists to solve the exploration problem defined by Eqs. (\ref{eq:Optim}, \ref{eq:C}) over a general graph $\mathcal{G}(V,E)$ within an arbitrary constant error $\mathrm{e}_r \in \R$ then  $\mathbf{P=NP}$. }
\vspace{10pt}

The proof of theorem \ref{thm:appxLongest}, given in Appendix \ref{sect:poofOfThm}, shows that the maximization of Eq. (\ref{eq:C}) over a graph $\mathcal{G}$ is more general than the Hamiltonian Path problem \cite{NetworksBook}. This proof leverages the work in \cite{Kaerger} which shows that a polynomial time algorithm which approximates the Hamiltonian Path within an arbitrary constant is equivalent to proving that $\mathbf{P=NP}$. Theorem \ref{thm:appxLongest} can be generalized to include a multitude of exploration and minimal convenience problems, such as that studied in \cite{Prentice}, and shows that optimality guarantees of algorithms such as that in \cite{Prentice} cannot be made. Theorem \ref{thm:appxLongest} provides theoretical limits of performance for exploration problems which rely on general graph structures produced by algorithms such as the PRM and implies that an exhaustive search must be used to maximize Eq. \eqref{eq:C}. 

\subsubsection{Optimization of the Reward} \label{sect:OptiMeth}
Given a set of LF feasible paths that satisfy the threshold in Def. (\ref{def:LocFeas}), the reward function can then be optimized to find the `best' exploration path via a graph search algorithm. A node $v_{\rm start}$ is added to $\mathcal{G}$ at the robot's initial pose estimate, and a goal vertex $v_{\rm goal}$ is defined as the centroid of one the members of $\mathcal{L}$; optimization of the location $v_{\rm goal}$ within $\mathcal{L}$ is left to future work.

Due to the non-linearity of expected entropy in the reward function in Eq. (\ref{eq:CSPLAM}) as well as the fact that the IE problem is maximizing a positive reward, additive graph search algorithms such as Dijkstra's and A* cannot be used \cite{LaValle2006}. This work utilizes a modified Depth-First Search (DFS) to iterate through all potential paths from $v_{\rm start}$ to $v_{\rm goal}$ to find the optimal path. 

The expected entropy computation in the reward function, Eq. (\ref{eq:CSPLAM}), increases intractably with the length of the measurement $Z_{t:T}$. In order to alleviate this scaling problem, the entropy of only the subset of cells which will be measured along the reference path $\hat{x}_{k_1:k_2}$ is computed. These measured cells are known due to assumption 4: $x_{t:T} \approx \hat{x}_{t:T}$. The measurements affecting a particular cell are denoted $Z^{c_i}$. Many of the cells in the exploration space have no relevant measurements and their expected change in entropy need not be computed, i.e.: $$\underset{Z_{t:T}}{\E}[H(C^i(T)] = H(C^i(t)).$$ 

\subsection{Reward Bound} \label{sect:Bound}
Computation of the reward in Eq. \eqref{eq:CSPLAM} can be greatly reduced by utilizing a tight bound on the entropy of cells. In the sequel derivation a single cell is analyzed. Thus the dependence of $Z$ on $c$ is dropped and the measurements are assumed to start at index 1. In order to obtain a bound for Eq. \eqref{eq:CSPLAM}, several mild assumptions are made. First, the probability of mis-detection, $1 - \theta$,  and false alarms, $1-\phi$,  are assumed identical. Second, each measurement, $Z_j$, is assumed to be taken from an $i.i.d.$ mixture of Bernoulli RVs. Finally, it is assumed that the sensor sampling frequency is fast in comparison to the motion of the robot, implying that some cells $C_i$ have a large number of samples, $n_z$. The Bernoulli distributions follow the probabilities:

\begin{equation} \label{eq:ProbMisDetect}
\begin{array}{ccc}
\theta:= P(Z_j = 1| c = 1) = P(Z_j=0| c=0) := \phi.  \\
\end{array}
\end{equation}

\noindent
Given the stated assumptions, Prop. \ref{prop:EntroBound} holds.

\prop{\label{prop:EntroBound} A bound on the reward of a cell: $\lim_{n \to \infty} \underset{Z_{t:T}}{\E}[H(C|\lbrace Z_{1},...,Z_{n_z} \rbrace)] \geq \bigg({1 \over 2} \log \Big({e \over 2} \Big) \bigg)$ } 

The proof of Prop. \eqref{prop:EntroBound} is left to appendix \ref{sect:proofOfEntroBound}. 

Note that the bound in Prop. \eqref{prop:EntroBound} does not depend on $\theta$, thus the assumption that $\phi = \theta$ is not restrictive. As such, the worst case (in terms of certainty) between the false alarm and mis-detection rates is taken to be $(1-\theta)$. In practice, this approximation can reduce the computation time of $R(X_{t:T})$ by orders of magnitude, which makes Prop. \ref{prop:EntroBound} a key result of this work. More formally, the computation required to evaluate the expected entropy of a cell traditionally takes $\BigO{K \times n_z}$ where $K \in \R$ \cite{Bishop06}. Proposition \eqref{prop:EntroBound} makes this computation a simple subtraction ($\BigO{1}$). In MATLAB, this difference in computation time is on the order of $10^3$ for $n_z = 100$. 

Using Prop. \ref{prop:EntroBound} the measured space is divided into two parts: that with many measurements $M_{\mathrm{cert}}$ , and that with few $M_{\mathrm{uncert}}$. Clearly $M(T) = M_{\mathrm{cert}}(T) \cup M_{\mathrm{uncert}}(T)$ and $M_{\mathrm{uncert}} \cap M_{\mathrm{cert}} =\O$ (i.e. these are `certain' and `uncertain' parts of the space).
\vspace{-.5em}
$$\sum_{C_i \in M_{\mathrm{cert}}}\underset{Z^{c_i}_{t:T}}{\E}[H(C_i)] \geq \mathcal{C}(M_{\mathrm{cert}}) \cdot \bigg({1 \over 2} \log \Big({e \over 2} \Big) \bigg) := -\bar{R}_{\mathrm{cert}}.$$

\noindent
 In practical scenarios such as those in the results section, a majority of the measured cells in $M(T)$ will also be in $M_{\mathrm{cert}}$(T), hence Prop. \ref{prop:EntroBound} significantly increases evaluation of Eq. \eqref{eq:CSPLAM}. The reward in Eq. (\ref{eq:CSPLAM}) is bounded by:
\vspace{-.25em}
\begin{align} \label{eq:Cappx}
\begin{split}
R(X_{t:T}) & \geq  \bar{R}(X_{t:T}) := \\
H(M(T)) & - \sum_{C_i \in M_{\mathrm{uncert}}} \underset{Z^{c_i}_{t:T}}{\E}[H(C_i)] + \bar{R}_{\mathrm{cert}}.
\end{split}
\end{align}

\subsection{Summary: The G-PIE Algorithm}
An exploration algorithm with localization constrains can now be fully described by combining the reward function in Eq. (\ref{eq:CSPLAM}), the constraint defined in Def. (\ref{def:LocFeas}), and Depth First Search (DFS): (Alg. \ref{Alg_GPIE}).

\setlength{\textfloatsep}{2pt}
\begin{algorithm}[h] 
$R_{\mathrm{best}} =0$\;
$X_{\mathrm{best}} = \emptyset$\;
$\mathcal{G}(V,E) = PRM(C_{\mathrm{free}},C_{\mathrm{obst}})$\;
$V \leftarrow Add \Big(v_{\mathrm{start}} = \hat{x}_0 \Big)$\;
$V \leftarrow Add \Big(v_{\mathrm{goal}} = x_{\mathrm{goal}} \Big)$\;
\For{$\big(X_{t:T}| \lb \bar{x}_0 = v_{\mathrm{start}}, \bar{x}_T = v_{\mathrm{goal}} \rb \big)$}{
	\vspace{6pt}
	\If{$P(X_T \in \mathcal{L}) \geq \alpha$}{
	$Compute: \bar{R}(X_{t:T});$
	
	    \vspace{6pt}
		\If{$\bar{R}(X_{t:T}) >R_{\mathrm{best}}$}{
		$R_{\mathrm{best}} = \bar{R}(X_{t:T})$ \\
		$X_{\mathrm{best}} = X_{t:T}$
		}
	}
}

\Return{$(X_{\mathrm{best}},C_{\mathrm{best}})$}
\caption{The Guaranteed Probabilistic Information Explorer (G-PIE) Algorithm. Depth First Search is used in line 6.} \label{Alg_GPIE}
\end{algorithm}

The Guaranteed Probabilistic Information Explorer (G-PIE) algorithm behaves as follows. First, the best reward and best path, $R_{\mathrm{best}}$ and $X_{\mathrm{best}}$, are reset, and a PRM graph, $\mathcal{G}$, is generated. Next, the start and goal nodes are added to $\mathcal{G}$, and each potential path from $v_{\mathrm{start}}$ to $v_{\mathrm{goal}}$ is checked for Localization Feasibility (LF). If a path, $X_{t:T}$, is feasible, its bounded reward, $\bar{R}(X_{t:T})$, is computed and checked against $R_{\mathrm{best}}$. If required, the best path and best reward are then updated, and the process repeats until all paths have been calculated or an allotted computational time has expired. Given sufficient computational time, the algorithm is guaranteed to return the optimal LF path within the current graph $\mathcal{G}$. In addition when using a complete variant of the PRM, the path found by G-PIE converges to the true optimal as $\mathcal{C}(V) \to \infty$ \cite{Frazzoli}.

\section{A Receding Horizon G-PIE} \label{sect:RHPIE}
The G-PIE algorithm is guaranteed to provide optimal exploratory trajectories. However, G-PIE relies on an exhaustive search, creating computational challenges, and does not take into consideration uncertainty of the robot's intermediate pose in the reward function. Theorem (\ref{thm:appxLongest}), shows that on a general graph structure, no exploration algorithms can guarantee being within an arbitrary constant of optimal. In addition, imposing an overly restrictive structure, such as a tree, on the $\mathcal{G}$ makes it poorly approximate optimal trajectories. These facts imply that any approximation algorithm should be focused on finding local optima. 

To address computational scaling, a receding horizon can be utilized. Note that as the horizon $T_1 \in \{1,2,3,... \}$ increases, this receding horizon approach will obtain the optimal path given by the G-PIE. Furthermore, a heuristic tail cost function is presented which balances information gain and pose uncertainty dynamically beyond the horizon $T_1$. This tail cost function provides better sub-optimal solutions and can be dynamically tuned to seek more conservative (LF) or exploratory paths.

\begin{figure}
\includegraphics[scale=.27]{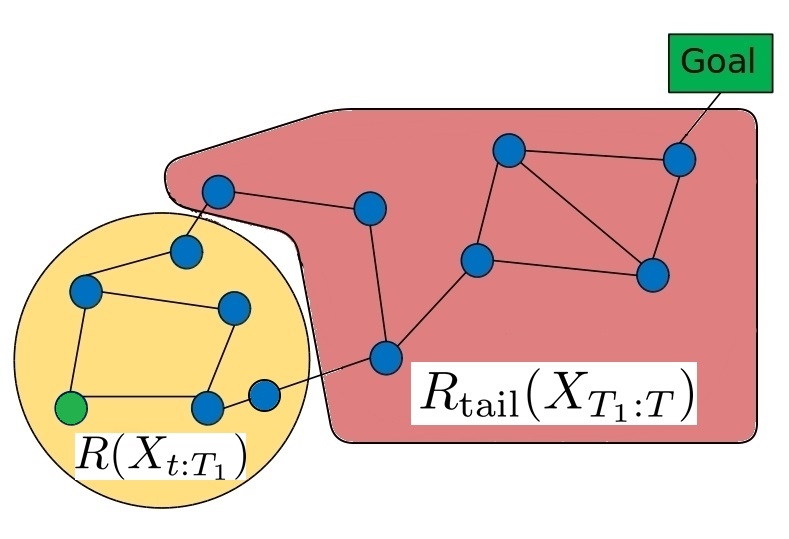}
\caption{An exemplification of the horizon $T_1 = 2$. Local paths are at most two nodes away from the green start node.}
\label{fig:TailCartoon}
\end{figure}

\subsection{An Augmented Reward Function}
The most common form of receding horizon control considers the optimization of a discrete time system over a finite horizon \cite{BertsekasBook}. In this paper, the reward function is optimized over a finite number of node visitations in the graph $\mathcal{G}$; exemplified in Fig. (\ref{fig:TailCartoon}). This is intuitive because the more nodes a robot visits, the longer its trajectory and  the associated time to complete this trajectory. 

The receding horizon reward function, $R_{\mathrm{rh}}(X_{t:T})$, is split into a local reward, $R(X_{t:T_1})$, and an estimated tail reward $R_{\mathrm{tail}}(X_{T_1:T})$. The use of $(t:T_1)$ is an abuse of notation since $T_1$ denotes an integer number of node visitations, as shown in Fig. \eqref{fig:TailCartoon}, while $t$ and $T$ are discrete time instances.
\begin{equation}
\label{eq:C_aug}
R_{\mathrm{rh}}(X_{t:T}) =  R(X_{t:T_1}) + R_{\mathrm{tail}}(X_{T_1:T})
\end{equation}

\noindent
If an exact estimate of $R_{\mathrm{tail}}(X_{T_1:T})$ were available \textit{a priori}, the optimal initial path, $X_{t:T_1} $, is recovered. In simple receding horizon control problems $R_{\mathrm{tail}}(\centerdot)$ is assumed to be zero or a large over estimate \cite{BertsekasBook}, but the appropriate selection of this tail reward function can lead to optimal or near optimal solutions \cite{pearl1984Heuristics}. 

The constraint in Def. (\ref{def:LocFeas}) must still be enforced by any returned path maximizing $R_{\mathrm{rh}}(\centerdot)$. Thus to ensure feasibility, it is prudent to consider a tail reward estimate which balances information gain and growth in pose uncertainty. To do this, the tail reward is defined as a scalarization of estimates of information gain, $B_{\mathrm{info}}(X_{T_1:T})$, and feasibility, $B_{\mathrm{pos}}(X_{T_1:T})$, with the scalarization parameter $\beta \in (0,1)$

\begin{equation}\label{eq:TailCost}
R_{\mathrm{tail}}(X_{T_1:T}) = \beta*B_{\mathrm{info}}(X_{T_1:T}) - (1-\beta)B_{\mathrm{pos}}(X_{T_1:T})
\end{equation}

\noindent
Here, $B_{\mathrm{pos}}(\centerdot)$  and $B_{\mathrm{info}}(\centerdot)$ are positive valued functions. The derivation of $B_{\mathrm{pos}}(\centerdot)$ and $B_{\mathrm{info}}(\centerdot)$ is presented next. Intuitively, when $\beta=0$, the returned path strives to make the constraint in Def. (\ref{def:LocFeas}) as non-binding as possible and attempts to ensure feasibility, but the optimization ignores the information content of $X_{T_1:T}$. Conversely, when $\beta = 1$, the optimizing path is more information rich, but may not satisfy Def. (\ref{def:LocFeas}). Since Def. \eqref{def:LocFeas} must be satisfied, this constraint is checked for all solutions and any solution not satisfying \eqref{def:LocFeas} is discarded. A similar formulation in a differing application is presented in \cite{Ferrari09}. 

The primary purpose of the cost in Eq. \eqref{eq:TailCost} is to find locally optimal yet \textit{feasible} tail trajectories. Thus, a small value of $\beta$ may be required to ensure the satisfaction of the probabilistic constraint. Conversely, it is desirable to set $\beta$ as large as possible to better approximate the true tail reward, and achieve better global performance. The trade between these goals, as $\beta$ varies, is studied in the results. 

Using Eq. \eqref{eq:TailCost} implies the optimization of $X_{T_1:T}$ is a shortest path problem. This is of vital importance because there are known algorithms which can solve this problem quickly, i.e. Dijkstra's. A judicious choice of $B_{\mathrm{pos}}(\centerdot)$, $B_{\mathrm{info}}(\centerdot)$, and $\beta$ is required to guarantee a path graph with positive edge weights which allows for the optimal use such powerful algorithms. 

To briefly summarize, the main goal in deriving $R_{\mathrm{tail}}$ is to achieve additivity and path independence in the tail sub-problem. This allows the use of shortest path algorithms to calculate $R_{\mathrm{tail}}$ and provides a large computational improvement over the full G-PIE solution. In particular, if the graph $\mathcal{G}$ has a maximum connectivity $\kappa$ and node count $\psi$, then the number of paths from one node to another node in $\mathcal{G}$ is at most $\BigO{\kappa^{\psi}}$, and each of these must be evaluated the the G-PIE. In contrast, the RH-PIE complexity becomes $\BigO{\kappa^{T_1}*\psi \log{\psi}}$, since each of the local $\BigO{\kappa^{T_1}}$ paths must be completed by using a Dijkstra like algorithm to calculate $R_{\mathrm{tail}}$. In addition, any path which maximizes $R_{\mathrm{rh}}$ in Eq. \eqref{eq:C_aug} must still satisfy the re-localization constraint. The two different terms in $R_{\mathrm{rh}}$ seek a compromise between safety, which must always be guaranteed, and information, whose global optimality is sacrificed by this approximate reward.

\subsection{Derivation of Uncertainty Penalty}
The uncertainty penalty, $B_{\mathrm{pos}}(\centerdot)$, penalizes growth in positional uncertainty and is the mechanism which enables the RH-PIE to find a feasible tail trajectory $X_{T_1:T}$. The penalty, $B_{\mathrm{pos}}(\centerdot)$, in Eq. (\ref{eq:TailCost}) must associate an uncertainty cost to traversing edges along the graph $\mathcal{G}$, and this penalty must be additive.  Second, $B_{\mathrm{pos}}$ must be path independent to allow use of shortest path algorithms. This means traversing an edge of $\mathcal{G}$ must incur the same penalty regardless of previous node visitations. In other words, the argument of $B_{\mathrm{pos}}$ reduces to a node ordering. In the case of robotic motion, the pose uncertainty of the robot is path dependent; the Curse of History \cite{pineau2003point}. Instead, this work uses a bound on the maximum growth in uncertainty when traversing an edge.

In \cite{Prentice} the authors assume an EKF is used for robot pose estimation. They show that, along a particular edge $e_i = (\hat{x}_{i}, \hat{x}_{i+1})$, the update equations can be simplified by using aggregate matrices; the observation matrix  $H^T Q^{-1}H \in \R^{n_x \times n_x}$, the noise matrix $L\in \R^{n_x \times n_x}$, and the propagation matrix $G\in \R^{n_x \times n_x}$. A detailed definition of these matrices cannot be given here and is given in \cite{Prentice}, while details on the EKF formulation are given in \cite{thrun2005probabilistic}. The update equation is now written as:
\begin{equation} \label{eq:BRMKalman}
\Sigma_{i+1} = L + G(\Sigma^{-1}_i +H^T Q^{-1}H)^{-1}G^T .
\end{equation}

\noindent
note that $i$ is a spacial rather than time index. This equation implies the system is observable (perhaps weakly) over an edge. Regardless of observability, the following analysis is still valid. The matrix $G$ is assumed to be invertible which is true in most problems \cite{Prentice}. 

\prop{ \label{prop:EigenBound} Suppose, $H^T Q^{-1}H$ and $\Sigma^{-1}_i$ are positive definite Hermitian matrices, then $\lambda_1(\Sigma_{i+1})$ is bounded by:

\begin{multline*}
\lambda_1 \Big(\Sigma_{i+1} \Big) \leq \lambda_1 \Big(L \Big)+ \\ \min \Big\{\lambda_1(G \Sigma_i G^T),\lambda_1(G(H^T Q^{-1}H)^{-1}G^T) \Big\}= \lambda_{\mathrm{bound}}.
\end{multline*}

}

\cor{ Using Prop. (\ref{prop:EigenBound}) the following holds: 

\begin{equation} \label{eq:DeltaEigenBound}
\begin{array}{rcl}
\Delta \Sigma(e_i)  & := & \Sigma_{i+1}-\Sigma_{i} \\
\lambda_1(\Delta \Sigma (e_i))& \leq & \lambda_{\mathrm{bound}} - \lambda_1(\Sigma_i)
\end{array}
\end{equation}
}

\noindent
The proof of Prop. (\ref{prop:EigenBound}) is given in Appendix \ref{sect:proofOfEigenBound}. A bound on the change in the largest eigenvalue of the robot's covariance matrix also bounds the trace of the covariance. Notice that this bound is still dependent on $\Sigma_i$. To make this bound path independent, notice that the argument inside $\min \{ \hspace{1pt}\centerdot \hspace{1pt}, \hspace{1pt} \centerdot \hspace{1pt}\}$ which is dependent on $\Sigma_i$ can be ignored and the inequality still holds. In other words:

\begin{multline}
\min \Big\{\lambda_1(G \Sigma_i G^T),\lambda_1(G(H^T Q^{-1}H)^{-1}G^T) \Big\} \leq \\ 
\lambda_1(G(H^T Q^{-1}H)^{-1}G^T).
\end{multline}

\noindent
The bound in Eq. (\ref{eq:DeltaEigenBound}) depends on $\Sigma_i$ through $\lambda_{\mathrm{bound}}$ and the subtraction of $\lambda_1(\Sigma_i)$. To eliminate this dependence, both of these terms must be addressed. Starting with $\lambda_{\mathrm{bound}}$, an analysis of Prop. (\ref{prop:EigenBound}) shows that the first term in the minimum takes into consideration the dynamics of the robot while the second term becomes infinite as the system becomes unobservable. Many authors assume the integral observability of the system \cite{Prentice}. Regardless, in scenarios where no pose information is available to the robot, it is practically necessary to maintain both terms in the minimum. One approach is to consider a worst case $\Sigma_{\mathrm{worst}}$ which has only one eigenvalue (i.e. symmetric uncertainty); the worst case could be the divergence of the pose estimate. Once $\Sigma_{\mathrm{worst}}$ is identified, this value can be used for all edges of $\mathcal{G}$ in lieu of $\Sigma_{i}$ in Prop. (\ref{prop:EigenBound}).

Thus, $\lambda_{\mathrm{bound}}$ has been made path independent, but Eq. (\ref{eq:DeltaEigenBound}) still depends on $\Sigma_i$. A looser approximation which requires no further assumptions is bounding Eq. (\ref{eq:DeltaEigenBound}) from above by $\lambda_{\mathrm{bound}}$ alone. This bound can be too loose, and it is therefore practical and convenient to define $\Sigma_{\mathrm{best}}$. In other words, defining $\Sigma_{\mathrm{best}}$ to be such that $\Sigma_k$ has a smaller spectral norm than $\Sigma_{\mathrm{best}}$ in all practical scenarios. $\Sigma_{\mathrm{best}}$ should be small enough that this level of uncertainty in position has little bearing on the performance of the robot in its mission. In this case, the most natural definition of $\lambda_1(\Sigma_{\mathrm{best}})$ is $ \gamma / \mathrm{dim}(X)$, where $\gamma$ is taken from Def. (\ref{def:localizable}).

Now that the bound in Eq. \ref{eq:DeltaEigenBound} is established, the pose uncertainty penalty associated with an edge is:

\begin{equation}
B_{\mathrm{pos}}(X_{T_1:T}) = \sum_{i=T_1} \lambda_1(\Delta \Sigma(e_i)).
\end{equation}

\noindent
Note that $B_{\mathrm{pos}}(\centerdot)$ is positive because the matrices involved are positive definite Hermitian. 

Interpreting this bound intuitively, notice that $L$ in Prop. (\ref{prop:EigenBound}) represents the uncertainty added due to process noise, $G \Sigma_k G^T$ is a transformation and scaling due to robot motion, and $G(H^TQ^{-1}H)^{-1}G^T$ is the net effect of expected measurements. Thus, $B_{\mathrm{pos}}$ increases due to robot motion and length of the reference path, and decreases with good pose observations along the path. 

\subsection{Derivation of Tail Information Reward}
In order for $R_{\mathrm{tail}}$ to be fully path independent, $B_{\mathrm{info}}$ must also be made path independent and additive. In reality, an edge $e_i$ can give more or less information based on how the robot moved prior to traversing $e_i$. Thus, an approximation must be made in order to achieve additivity and path independence. 

The key challenge in the approximation of information gain is due to information `overlap'. If the robot traverses $e_i$ before $e_j$ more information is gathered along $e_i$ in comparison to the case when the robot first traverses $e_j$ then $e_i$. Three approximations are presented, which include an under bound, over bound, and average. All three use the fundamental idea of dividing the cells of the exploration space into regions where each edge $e_i$ has an associated set of cells. The cells associated with $e_i$ are then assumed to be independent of traversal along $e_j$. This assumption breaks the path dependence while also drastically improving computation speed. 

The first approach is the simplest and assumes that any cells within sensor range of $e_i$ are not affected by the traversal of any other edge. This implies that even if a cell $c_j$ is within sensor range of two or more edges, the reward for traversing $e_i$ is the same as if none of the cells within its range are observed before traveling along $e_i$. Let $M_{e_i}$ be the portion of the exploration space visible by traversing edge $e_i$, then the tail information reward can be expressed as:

\begin{equation}\label{eq:EdgeOverApx}
B^{\mathrm{over}}_{\mathrm{info}}(X_{T_1:T}) = \sum_{e_i \in X_{T_1:T} } \Big[ H(M_{e_i}) - \E_{Z_{e_i}}[H(M_{e_i})] \Big].
\end{equation}

\noindent
This value is really an over bound of the expected information gain along $e_i$. This method is fast to compute, and gives a reasonable estimate when the optimal path does not traverse the same area many times (loops).  If the optimal path does loop this method will `double-count' information in nearby edges. 

The second method partitions the exploration space into Voronoi regions about each edge $e_i$ \cite{LaValle2006}. Once this is accomplished, any cell whose centroid is in the Voronoi region of $e_i$ is associated with $e_i$. Let $M^{\mathrm{under}}_{e_i}$ be the portion of the exploration space $M$ associated with edge $e_i$ through the Voroni regions. The tail information reward is:

\begin{equation}\label{eq:edgeUnderApx}
B^{\mathrm{under}}_{\mathrm{info}}(X_{T_1:T}) = \sum_{e_i \in X_{T_1:T} } \Big[ H(M^{\mathrm{under}}_{e_i}) - \E_{Z_{e_i}}[H(M^{\mathrm{under}}_{e_i})] \Big].
\end{equation} 

\noindent
This is a fast under-bound of the information gained from traversing an edge, but it may give large under estimates to a subset of edges. 

A final technique attempts to compensate for the weaknesses of the first two methods. In this approximation, the parts of the exploration space which are in range of several edges are penalized, but all cells in the sensor range of $e_i$ are associated with $e_i$. 

\begin{equation} \label{eq:apxCompromise}
B^{\mathrm{ave}}_{\mathrm{info}}(X_{T_1:T}) = \sum_{e_i \in X_{T_1:T} } \Bigg[ H(M_{e_i}) - \E_{Z_{e_i}} \Big[\sum_{c_j \in M_{e_i}} c_j/k_j \Big] \Bigg]
\end{equation}

\noindent
where $k_j$ is the number of edges which have $c_j$ in their sensor range. The value in Eq. (\ref{eq:apxCompromise}) is neither an under nor over estimate of the information gained along $e_i$. To see this, consider the case when all edges within range of $c_j$ are traversed by a single path. The concavity of the expected entropy causes an over estimate by Eq. (\ref{eq:apxCompromise}). Conversely, if a path only traverses a single edge affecting $c_j$ and $k_j>1$, Eq. (\ref{eq:apxCompromise}) produces an underestimate. 

\subsection{The importance of $\beta$}
Given the definitions of $B_{\mathrm{pos}}$ and $B_{\mathrm{info}}$, the selection of $\beta$ must be considered. The form of $R_{\mathrm{tail}}$ shown in Eq. (\ref{eq:TailCost}) transforms the problem of finding $X_{T_1:T}$ into a shortest path problem if $\beta$ is selected such that there are no negative loops in the graph $\mathcal{G}$. Since $B_{\mathrm{pos}}$ is strictly positive and $B_{\mathrm{info}}$ is non negative, such a $\beta$ can always be found. In addition, $\beta$ controls how much weight is given to information gain versus maintaining low pose uncertainty. 

Recall that, in an un-directed graph $\mathcal{G}$, any negative edge results in a negative  loop between two nodes. Thus, $\beta$ must be set to ensure non-negative edge weights throughout the entire graph, which in turn enables shortest path algorithms to return an optimal path. Although this requirement is sometimes ignored, as in \cite{Ferrari09}, it is vital to ensure optimality in the tail sub-problem. Note that even when $\beta$ is larger than the value which ensures negative weights, Dijkstra-like algorithms can return sub optimal paths. Note also that by setting $\beta$ high, returned paths can be much more information rich even if sub optimal for that particular value of $\beta$.  Once $\beta$ is set $X_{T_1:T}$ can be determined. The Receding Horizon Probabilistic Information Explorer  (RH-PIE) can be fully described in Alg.(2).

\begin{algorithm} \label{alg:RH-PIE}
\caption{RH-PIE}
$R_{\mathrm{best}} =0$\;
$X_{\mathrm{best}} = \emptyset$\;
$\mathcal{G}(V,E) = PRM(C_{\mathrm{free}},m)$\;
$Add \Big(v_{\mathrm{start}} = \hat{x}_0 \Big)$\;
$Add \Big(v_{\mathrm{goal}} = x_{\mathrm{goal}} \Big)$\;
$\beta = \beta_{\mathrm{user}}$

\For{$\big(X_{0:T_1} \in \mathcal{X}_{0:T_1}| \lb \bar{x}_0 = v_{\mathrm{start}} \rb \big)$}{
    $[X_{T_1:T}, R_{\mathrm{tail}}(X_{T_1:T})] = ShortestPath(\hat{x}_{T_1}, \bar{x}_T)$
    
	\If{$P(X_T \in \mathcal{L}) \geq \alpha$}{
	$Compute: R_{\mathrm{rh}}(X_{0:T})= R(X_{0:T_1}) +  R_{\mathrm{tail}}(X_{T_1:T})$\;
	    \vspace{8pt}
		\If{$R_{\mathrm{rh}}(X_{0:T}) >R_{\mathrm{best}}$}{
		$R_{\mathrm{best}} = R_{\mathrm{rh}}(X_{0:T})$ \\
		$X_{\mathrm{best}} = X_{0:T}$
		}
	}
}

\Return{$(X_{\mathrm{best}},R_{\mathrm{best}})$}

\end{algorithm}
\noindent
The RH-PIE algorithm is similar in structure to the G-PIE algorithm. The primary difference is that, $R_{\mathrm{rh}}$ is calculated for the set of local paths within the user defined horizon, $T_1$.

\section{SIMULATION AND EXPERIMENTAL RESULTS} \label{sect:Results}
In order to fully understand both the theoretical and practical behavior of the G-PIE and RH-PIE algorithms, two sets of results are presented. 

\begin{figure}
\includegraphics[scale=.6]{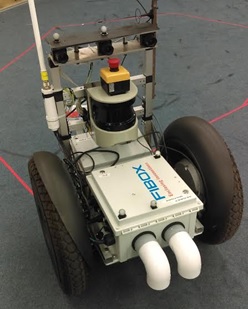} \includegraphics[scale=.52]{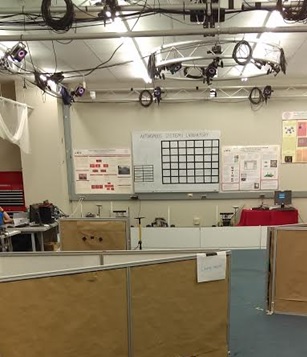}
\caption{Shown on the left is the Segway RMP50 robotic platform. It has GPS, odometry, LIDAR (LMS511), IMU, and camera sensing capabilities. The right image shows the experimental obstacles and VICON positioning system used to provide ground truth.}
%\vspace{-.5em}
\label{segway}
\end{figure}

\subsection{Models Sensors and Experimental Setup}
In both the simulation and experimental results, the robot is modeled as a unicycle with direct velocity/turn-rate control $(v,\omega)$ \cite{thrun2005probabilistic}. In order to match the modeling assumptions made between the simulations and experiments, a discrete-time, first order, linear, input-output dynamics model was fit to data from the Segway platform. A kinematic state feedback controller is used for trajectory generation and tracking \cite{klanvcar2007tracking}. This model is also used to inform the predictive step in the RH-PIE algorithm as well as in an EKF filter, which provides pose estimation.

Location measurements are noisy relative range and bearing to known landmarks (implicitly generating LPAs). Areas of interest are represented as a grid, and noisy binary measurements (interesting, uninteresting) are taken at 10Hz. In these scenarios, the edges of objects are taken to be `interesting'. Thus, in the experiments, measurements are returns from a SICK LMS511 LIDAR. The measurements' speed implies enough information measurements can be expected to form a large subset of cells in $M_{\rm{cert}}$ in Eq. (\ref{eq:Cappx}), assuming a robot speed of 1-3 m/s. The high number of measurements implies Prop. \ref{prop:EntroBound} is instrumental in accelerating the computation of the path reward by replacing Eq. (\ref{eq:CSPLAM}) with Eq. (\ref{eq:Cappx}).

In the experiments, ground truth of the robot pose is obtained using a VICON motion capture camera suite. In addition, VICON also allows the generation of software-based point landmarks, which are the basis of relative range and relative bearing pose measurements. Independent white Gaussian noise is added to each measurement. In this way, the measurement within LPAs can be precisely controlled and matched to modeling assumptions. The SICK has a 190\degree   field of view and its range is restricted to 2m due to the constrained laboratory environment. Note that the robot does not know \textit{a priori} that obstacle boundaries are interesting. 

Comparing the experimental and simulation setups, modeling errors in robot motion, as well as non-whiteness and non-Gaussianity of the realized noise are the only modeling differences. The practical differences between experimental and simulation settings include the necessary use of a reactionary obstacle avoidance procedure, and imperfectly synchronized measurements of the environment and robot position.

\begin{figure}
\centering
\includegraphics[scale=.28]{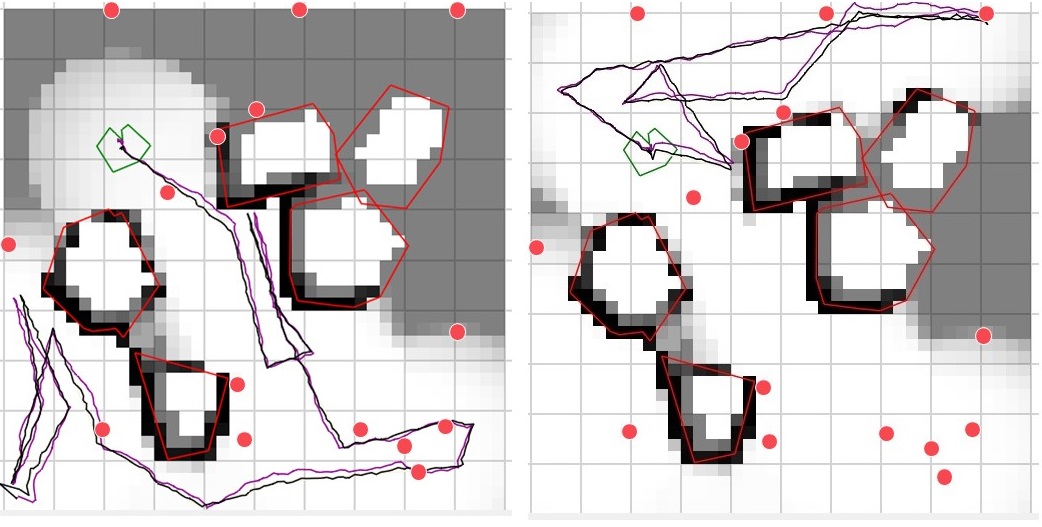} 
\caption{The robot starts at the bottom left with no information about the areas of interest and explores while traveling to $\mathcal{L}$. Landmarks are denoted in red, and $\mathcal{L}$ is composed of a single LRA in green. The robot's estimated EKF path is in black while its realized path is in purple. After a G-PIE path, the robot has discovered some areas of interest but a large area remains unexplored. After a second G-PIE path, the robot has discovered almost all areas of interest with high confidence.}
%\vspace{-.5em}
\label{fig:map2}
\end{figure}

\subsection{Simulation Results}

Three sets of simulations are presented to demonstrate the effectiveness of the G-PIE algorithm, verify claims, and validate assumptions. First, a qualitative discussion of the G-PIE algorithm's behavior is given. This demonstrates its functionality and its ability to repeatedly plan paths that enable long term autonomy. Second, the convergence of expected entropy to the  bound in Prop. \ref{prop:EntroBound} is analyzed and the bound's assumptions are scrutinized. Finally, Monte Carlo simulation results are shown comparing expected and realized constraint satisfaction and information gain. 

Figure (\ref{fig:map2}) shows an instance of two iterations of the G-PIE algorithm, exemplifying the trade between information gain and localization. Here, areas of interest are assumed to be the edges of objects, but the robot does not know this \textit{a priori} and cannot sense through obstacles. The discovered areas of interest are colored in black, while still unknown regions are in gray. The robot starts in the bottom left corner of the map. In the first iteration, Fig. (\ref{fig:map2}) left, the G-PIE algorithm selects the path with the highest possible information gain while guaranteeing a success of 95\% probability of terminating in $\mathcal{L}$. The robot navigates the environment, collecting information about areas of interest. The robot then terminates within the LRA, and its pose estimate becomes more confident due to location rich measurements. The updated pose enables the robot to continue to explore as shown in Fig. (\ref{fig:map2}) right, where the G-PIE algorithm replans from its current location back to $\mathcal{L}$. After two iterations the estimate of the regions of interest in Fig. (\ref{fig:map2}) right is produced. In addition, the user defined threshold of $\alpha = .95$ ensures that each iteration has a 95\% confidence of path completion. The remaining unexplored area is too risky, primarily because there are no nearby landmarks with which to localize. 

Because $\mathcal{L}$ is only one area, the G-PIE algorithm generates paths where the robot moves around unknown areas and loops back to the same region ($\mathcal{L}$). In the case where $\mathcal{L}$ is composed of multiple LRAs, the selection of the appropriate LRA can be added to the optimization.

\subsubsection{Convergence of the Expected Entropy }
\begin{figure}
\centering
\includegraphics[scale=.25]{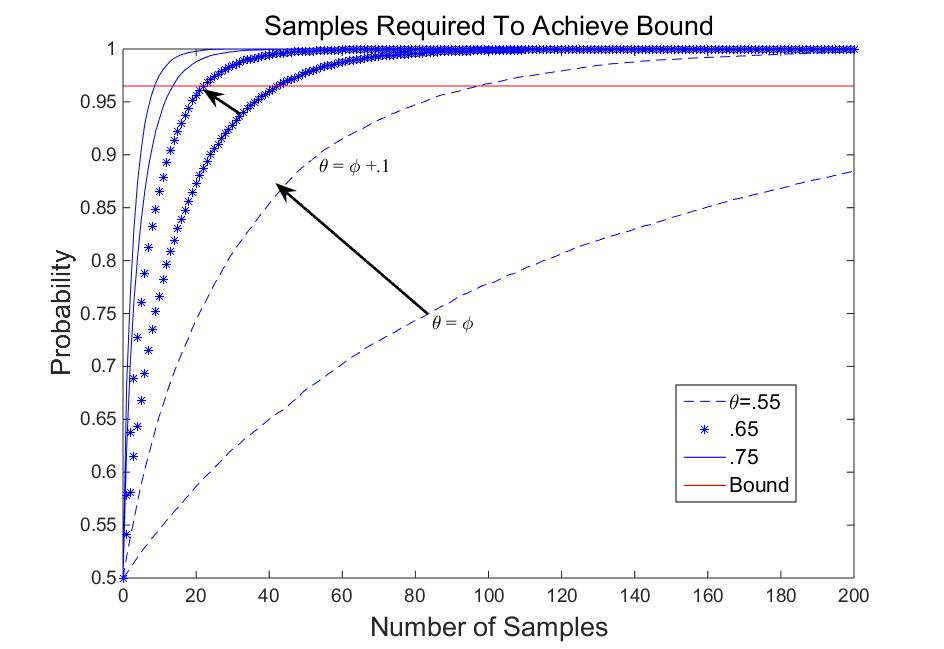}
\caption{Expected probability of interest, $p(C_i=1|Z_{1:n})$,of a cell as a function of the number of expected samples ,$n$, for three values of $\theta$. Curves with corresponding styles show the difference between the cases where $\theta = \phi$ and $\theta = \phi +.1$.}
\label{fig:symetric}
\end{figure}

In the development of Prop. \ref{prop:EntroBound}, the mis-detection and false alarm rates, $(1-\phi)$ and  $(1-\theta)$, are assumed to be identical; this section studies this assumption and the applicability of the entropy bound in Eq. (\ref{eq:Cappx}). Figure (\ref{fig:symetric}) plots $p(c=1|Z^c_{1:n})$ from the bound in Eq. (\ref{prop:EntroBound}) versus number of expected measurements, $n$, of that environment cell. A total of 10,000 Monte-Carlo cases were run using pseudo random numbers to generate samples of measurements from $p(Z^c|c=1)$ for a $3 \times 3$ test set of three cases of mis-detection/false alarm: $\theta = \phi$; $\theta = \phi +.1 $; $\theta = \phi -.1$. Three cases of $\theta$ are considered, and the prior distribution of a cell being interesting is assumed to be uniform 50\%. Figure (\ref{fig:symetric}) shows the symmetric case $(\theta = \phi)$ converges within 15 samples for mis-detection detection rates of 25\% $(\theta = 75\%)$, which is representative of sensors in real operating environments \cite{VisTracking_Dollar}. This result is equivalent to 1.5 seconds of observation with sensors operating at 10Hz, which is realistic in practice. At $\theta = 55\%$, the convergence is much slower ($>200$ samples). At this rate of false alarms, the sensor returns are incorrect $45\%$ of the time (nearly a coin flip). Regardless, the bound is still met near 400 samples (not shown due to scale), or 40 seconds of observation at 10Hz. Thus, the robot must be in sight of a particular grid cell for reasonable period, even with a poor sensor, for Prop. \ref{prop:EntroBound} to be valid. 

Figure (\ref{fig:symetric}) shows that, for the non-symmetric cases $(\theta \neq \phi)$, convergence is faster than the worst case scenario of $\theta = \phi \in [.55, .65,.75]$, respectively. This convergence trend is consistent for any values of $\theta$ and $\phi$. As a result, Fig. (\ref{fig:symetric}) verifies that the assumption $\theta = \phi$ is conservative. 

The G-PIE must evaluate hundreds or thousands of potential paths. Thus, the computation of the reward becomes significant. In  MATLAB on an I5 Intel processor, computing the expected entropy of 1000 cells with 30 expected samples takes $\sim 1 \mathrm{sec}$ of computation.  Using Prop. \ref{prop:EntroBound} instead requires  ${10^{-3}}$ times less computation. The study shown in Fig. (\ref{fig:symetric}) gives a guide for the number of samples required to use Prop. \ref{prop:EntroBound}. The number of required samples is the crossing point between the expected entropy of a cell and the derived bound for a given $\theta$ and $\phi$. Any number samples greater than this crossing point assures that Eq. (\ref{eq:Cappx}) is valid.
 
\subsubsection{Difference in Expected Path Completion and Exploration v.s. Realization}
A Monte Carlo (MC) study is performed to evaluate Desired versus Predicted versus Realized achievement of the G-PIE planner. The Desired Achievement is simply $\alpha$ (Def. \ref{def:LocFeas}), and determines the ability of the robot to terminate inside $\mathcal{L}$. The Predicted Achievement is the probability of paths returned by the G-PIE algorithm terminating inside $\mathcal{L}$: this must always be above $\alpha$ by construction. The Realized Achievement is the proportion of MC simulated runs which terminated inside $\mathcal{L}$. Obstacles, LPA, and LRA regions are randomly generated at each run. A failure is a path which does not terminate inside $\mathcal{L}$, or a collision with obstacles due to poor localization. By varying the given threshold $\alpha$ from .5 to .9, Fig. (\ref{fig:AvgFeas}) is obtained. Fig. (\ref{fig:AvgFeas}) plots Predicted (solid blue), and Realized (dashed red) Achievement as a function of Desired Achievement (solid,shaded black). Each data point corresponds to 10,000 MC runs. While $\alpha$ varies from .5 to .9, the MC Predicted Achievement correspondingly varies from  .86 to .94. This implies that the robot is able to find informative paths that also have a high probability of termination in $\mathcal{L}$ (i.e. the Predicted Achievement $>> \alpha$). The Realized Achievement lags behind Predicted  Achievement by approximately 2\% in each case due to obstacle collisions and non-linearity of the robotic system. Note that the G-PIE algorithm always ensures a higher path completion/localization than the given threshold $\alpha$ (i.e. G-PIE is conservative from the user's perspective).

\subsection{Experimental Results of RH-PIE}

In order to verify the practicality of the RH-PIE algorithm as well as analyze the effects of algorithmic parameters and environmental complexity, several sets of hardware experiments are presented. In this section, three key questions are addressed: 1) What effect does the choice of $\beta$ have on the solutions generated by the RH-PIE algorithm and how restrictive is the positive edge weight requirement for tail cost optimality? 2) What is the effect of the theoretical assumptions, particularly environmental complexity in the RH-PIE algorithm's development? 3) What effect does the look ahead distance have on path quality, optimality, and computational complexity?

In these experiments, three distinct maps are used. The choice of these environments enables the study of key parameters independently: obstacle complexity, initial position, look ahead distance, and $\beta$. In each map, two LRAs are provided in the same locations, shown in Fig. (\ref{fig:BetaExamples}) as green polygons.  The map is $3.5 \textrm{m} \times 9.5 \textrm{m}$ and has six localization landmarks as seen in Fig. \eqref{fig:BetaExamples}. It is important to note that each trial utilized the same underlying graph, $\mathcal{G}$, of 80 nodes to plan over. The graph is generated with a minimum connection distance of $0.5$m and a maximum connection distance of 1m. This implies that a look ahead of 1 is between 0.5m and 1m. The location and quantity of positional landmarks remains the same between maps. In terms of obstacles, the first map contains no obstacles, the second map contains one large central obstacle, and the third map contains three obstacles. Finally, to ensure repeatability, an automated initialization procedure was implemented to ensure the robot began each trial within 5cm and $1 \degree$ of its intended initial condition (IC).  

\subsubsection{The effect of $\beta$}

Recall that when $\beta$ is near $0$, the RH-PIE algorithm seeks only conservative information gathering paths which maintain an accurate pose estimate away from the local neighborhood of the robot. Conversely, setting $\beta$ near $1$ should cause the robot to exhibit more exploratory behavior. Regardless, the RH-PIE algorithm must satisfy the localization constraint and thus must guarantee relocalizing with high probability.

To evaluate the real effect of the $\beta$ parameter on $R_{\mathrm{tail}}$, trials were run on the three proposed maps using three different initial conditions (ICs). The parameter $\beta$ is swept from .1 to .9 in increments of .2. Thirty trials per map are performed for a total of 90 trials. The only variables modified are the IC and $\beta$. The robot was given a myopic look-ahead distance of $T_1 = 1$, 0.5-1m, while the feasibility threshold was maintained at $\alpha = .95$. It is noted that when $\beta>.2$, positive edge weights on $\mathcal{G}$ are not maintained and the path returned is not guaranteed to be optimal in terms of $R_{\mathrm{tail}}$. Despite the loss of tail optimality, the variation in $\beta$ above the .2 threshold yields qualitatively interesting and intuitive behavior.

\begin{figure}
\centering
\subfloat[No Obstacle Map]{\includegraphics[scale=.55]{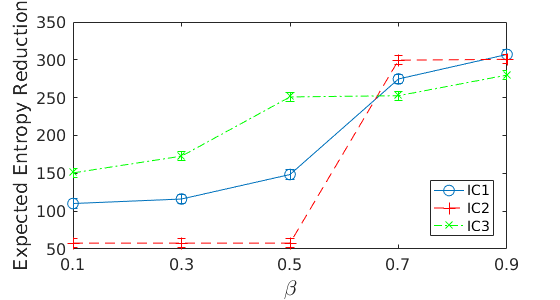}}

\centering
\subfloat[One Obstacle Map]{\includegraphics[scale=.55]{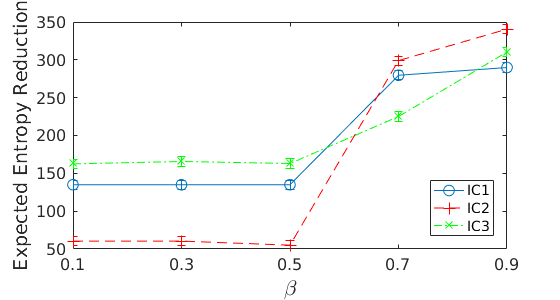}}

\centering
\subfloat[Three Obstacle Map]{\includegraphics[scale=.55]{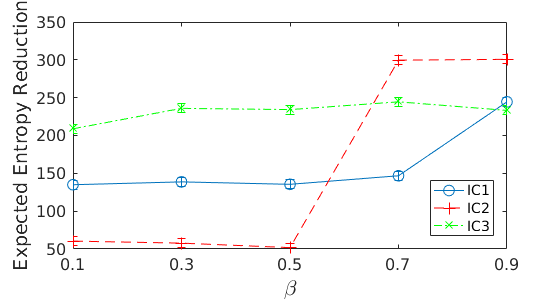}}

\caption{The expected information gain as a function of $\beta$ for three environments and three ICs. For  $\beta < .02$, the $R_{\mathrm{tail}}(\centerdot)$ is guaranteed to be optimal. Any returned path satisfies $\alpha = .95$. The maximum error due to IC variation is denoted by error bars.}
\label{fig:BetaSweep}
\end{figure}

Fig. (\ref{fig:BetaSweep}) plots Expected Entropy Reduction (information gain) as a function of $\beta$. Clearly, as $\beta$ increases the expected entropy reduction produced by $X_{\mathrm{best}}$ trends upward. This is because more weight is given to exploratory behavior. At values of $\beta$ which ensure tail reward optimality ($\beta \leq .2$), the algorithm is relatively insensitive to changes in $\beta$. For very small values of $\beta$, the RH-PIE algorithm exhibits localization seeking behavior and strives to observe as many landmarks as possible at the expense of distance traveled. Note the large jump in expected entropy reduction between $\beta = .5$ and $\beta = .7$. This is due to local edge weights around the robot's IC becoming negative and allowing more global exploration. Initial Condition 2 has the largest jump because the robot starts off near the goal, and the myopic look-ahead of 1 does not allow much exploratory action. Thus the robot must rely almost entirely on  $\beta$ to allow it to explore away from the goal location. 

\begin{figure}
\centering
\subfloat[$\beta = 0.1$]{\includegraphics[scale=.448]{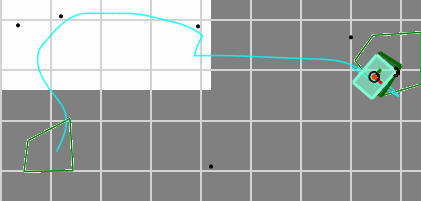}} \label{Beta1}
\centering
\subfloat[$\beta = 0.7$]{\includegraphics[scale=.448]{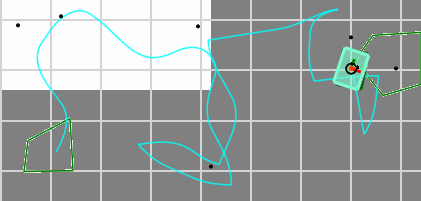}}\label{Beta5}
\caption{Two RH-PIE paths for two values of $\beta$. The robot is attempting to traverse the cyan path starting in the top right hand corner, to the bottom left LRA (green polygon). The true robot position is shown as a translucent cyan, while EKF estimate is translucent dark green.  A white cell implies nothing of interest exists, while black indicates a near certainty of an area of interest. The localization landmarks are denoted as black circles. }
\label{fig:BetaExamples}
\end{figure}

Fig. (\ref{fig:BetaExamples}) shows two different planned paths on the obstacle free map for $\beta =.1$ and $\beta =.7$. The robot begins in the top right corner of the map near an LRA and plans using the RH-PIE. The short look ahead helps to reveal the effects of $\beta$ more explicitly. The robot plans to the bottom left LRA while exploring. The absence of the EKF estimate implies that the EKF estimate and truth are nearly identical. Note the prior distribution imposed on areas of interest: the top left corner of the map is known to contain nothing of interest \textit{a priori}. Thus, the only reason that the robot should traverse this region while planning a path to the bottom left LRA is to observe the cluster of positional landmarks (black dots). It is evident that the robot exhibits strong localization seeking behavior when $\beta = .1$, and chooses to pass through the uninteresting area to see the positional landmarks. In contrast, when $\beta = .7$ the robot circles the two intermediate localization landmarks to ensure a feasible trajectory, but creates a highly exploratory trajectoy.

It is also important to note how environmental complexity and $\beta$ interplay. Figure \eqref{fig:BetaSweep} shows that the trends in entropy reduction remain relatively unchanged as the number of obstacles grows. However there are some differences in expected entropy reduction which result from the fact that obstacles generate different homotopy classes. This difference is especially evident when the exploration space is limited to `corridors' generated by obstacles, as is the case in the three obstacle map. In Fig. (\ref{fig:BetaSweep}a) and Fig. (\ref{fig:BetaSweep}c) for IC3, such an effect is clear. In the three obstacle case, the robot is forced into an exceptionaly exploratory corridor even for small $\beta$. Thus, expected information gain is higher in the three obstacle case as opposed to the obstcle free case for $\beta = 0.1$, even though the obstacle free case allows for more path flexibility. Conversly, the trend for IC3 in the three obstalcle map remains relatively flat because the robot has little else explore in its limited map.

\begin{figure}
\centering
\includegraphics[scale=.6]{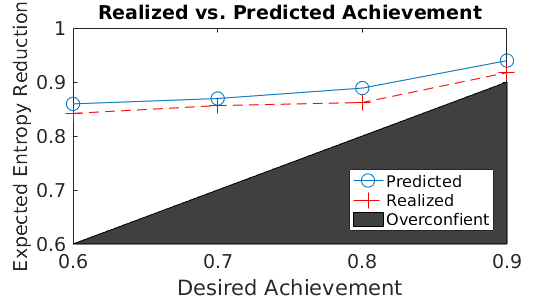}
\caption{ The predicted success rate for goal achievement as compared to simulated Monte Carlo realizations.}
\label{fig:AvgFeas}
\end{figure}

\subsubsection{The effect of environmental complexity on theoretical guarantees}
In order to analyze the effect of environmental complexity (existence and density of obstacles), a minimum of 30 trials were run on each of the three obstacle scenarios (maps). The robot begins exploration at the bottom center of the map, and attempts to explore the entire space. The robot is required to terminate inside a randomly chosen LRA with $\alpha \geq .95$. Landmark density was the same in each scenario, and relatively dense given the constrained laboratory environment. Attaining the LRA is considered a success, while encountering an obstacle, filter inconsistency, and an inability to attain the LRA are considered failures. The robot uses a rudimentary obstacle avoidance procedure: to avoid collisions, operation is ceased and a path is replanned if an obstacle is detected in the immediate path. Such obstacle detections are considered failures because the robot can attain the LRA by luck after many obstacle detections and subsequent re-planning stages.

The robot only replans if: 1) it reaches the LRA 2) it encounters an obstacle (due to an inaccurate pose estimate) 3) its EKF estimate becomes inconsistent and it fails to re-localize 4) it completes its planned path, but does not reach the LRA. Each re-planning stage utilizes an updated posterior estimate of the exploration space based on measurements taken by the SICK.

\begin{figure}
\centering
\includegraphics[scale=.45]{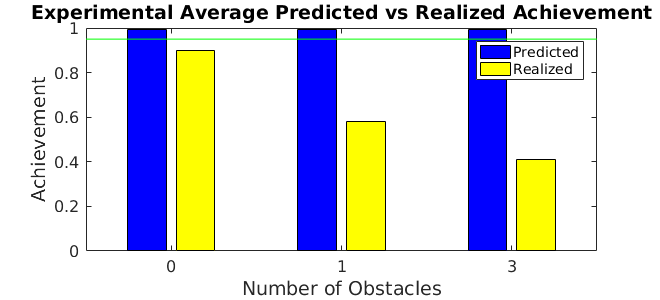}
\caption{ This figure shows the difference between Desired and Realized Achievement for three different obstacle scenarios.}
\label{fig:ExperimentalAchievement}
\end{figure}

Figure (\ref{fig:ExperimentalAchievement}) shows the results of these experiments. The robot's Desired Achievement is shown by the solid green line. The blue bars in Fig. (\ref{fig:ExperimentalAchievement}) show Predicted Achievement of just over .98 in each case. The scenarios' setup and the result in Fig. \eqref{fig:ExperimentalAchievement}, imply that paths of similar high Predicted Achievement were found between scenarios. The Realized Achievement from 30 runs is shown in yellow.

In the no obstacle case, three failures were observed where one failure was due to the robot exiting the defined map. This failure can be seen as an obstacle collision (the obstacle here being the map boundary). Controlling for this event, the realized achievement is $\approx .93$. Comparing this result with the simulation results in Fig. \eqref{fig:AvgFeas} provides valuable insight. In the simulation environment, maps were generated randomly and were more complex, in obstacle number and area coverage, than any of the experimental setups. In addition, the simulated robot, while following the same motion and measurement models, had no obstacle avoidance procedure. Thus, just as in the experimental procedure, any obstacle collisions were deemed failures. Whereas the simulations were conservative in complex environments, the experiment shows that the effects of model mismatch and un-modeled sensor noise made the constraint imposed the G-PIE and RH-PIE slightly optimistic. 

% Regardless, this small difference is still statistically significant at ($p = .05$). Therefore, in the benign no obstacle case, the RH-PIE qualitatively behaves as the theory predicts albeit with a statistically significant difference in performance. 

A dramatic drop off in performance is seen in the obstacle scenarios in Fig. \ref{fig:ExperimentalAchievement}. The analysis of the no obstacle case and experimental data imply that this drop off in performance is due to poor intermediate localization which triggered the obstacle avoidance procedure. Thus, as a practical point, the performance of the G-PIE and RH-PIE is not agnositic to the particular implementation of obstacle avoidance. In addition, this result shows that, in contrast to claims made in \cite{Prentice,agha2011firm}, path completion is not always dominated by terminal covariance. This dominance assumption can be fragile when sparse localization measurements are available for a non-linear system. 

The fact that path completion can be dominated by intermittent positional error is a valuable finding. Again, in simulation, the RH-PIE and G-PIE algorithms are able to attain conservatism in more complex senarios, as shown in Fig. (\ref{fig:AvgFeas}). This finding is consistent with simulation results provided in \cite{Prentice, agha2011firm}. Conversely, the experimental data suggest that metrics based on path completion, such as those presented here and in \cite{Prentice, agha2011firm}, are not sufficient to ensure conservatism in real world scenarios.  In particular, such path completion metrics should be augmented with direct consideration of intermediate localization. In addition, path completion metrics' interactions with obstacle avoidance procedures should be studied if they are to be practically useful and provide conservative results. 

\subsubsection{The effect of look ahead distance on path reward}

\begin{figure}

\centering
\subfloat[][$T_1=1$]{\includegraphics[scale=.6]{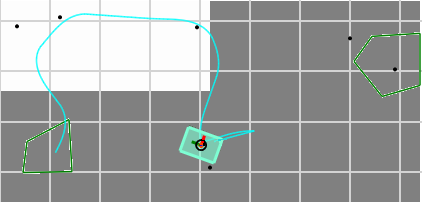}}

\centering
\subfloat[][$T_1=3$]{\includegraphics[scale=.6]{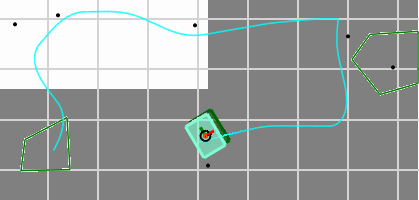}}

\caption{Two cases of the same obstacle scenario with the same initial conditions (IC1), but differing time horizons $T_1$. The robot is attempting to traverse the cyan path starting in the bottom center, and planning to the bottom left Localization Rich Area. $\beta = 0.1$.}
\label{fig:LookExamples}
\end{figure}

To analyze the effect of look ahead distance on performance, a set of experiments was conducted. The robot starts in three different initial conditions in the same three exemplary maps, and $\beta = 0.1$ is constant; thus any change in expected entropy reduction is solely due to the change in $T_1$. The look ahead distance varied from 1 to 3 nodes away from the starting location (0.5-3m). Each large grid cell in Fig. \ref{fig:LookExamples} is $1 \mathrm{m}^2$. Each data point is the average of 3 trials: a total of 81 trials were performed. In this case any variability within a trial set is due to inevitable variation in the exact initial position. 

\begin{figure}

\centering
\subfloat[Obstacle Free Map]{\includegraphics[scale=.55]{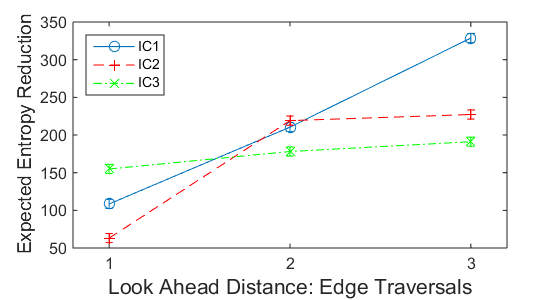}}

\centering
\subfloat[One Obstacle Map]{\includegraphics[scale=.55]{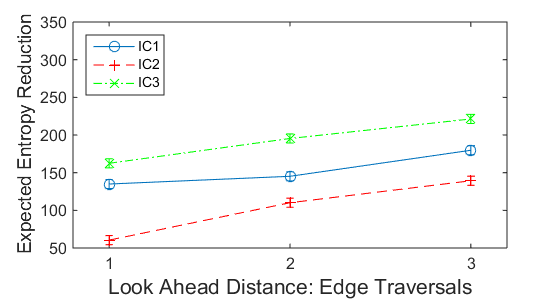}}

\centering
\subfloat[Three Obstacle Map]{\includegraphics[scale=.55]{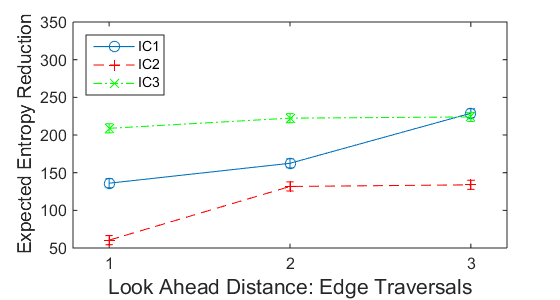}}
\caption{This set of figures shows the growth in expected information gain as a function of increasing look ahead distance. The parameter $\beta= 0.1$ is constant. The maximum error due to IC variation is denoted by error bars.}
\label{fig:LookSweep}
\end{figure}

Figure (\ref{fig:LookSweep}) plots expected entropy reduction as a function of $T_1$. As expected, an increase in the look ahead distance of the RH-PIE algorithm results in increased expected entropy reduction monotonically in all cases. In each map and set of ICs, there are two rates of increase: slow increases as seen in the single obstacle case, and more dramatic increases as seen in the obstacle free and three obstacle case. To understand these rates more clearly, consider Fig. (\ref{fig:LookExamples}), which shows the same scenario for two different horizon lengths. With a single step look ahead policy, the robot fails to take into consideration the localization landmark at the top right hand corner and subsequently takes a less exploratory path. 
With a two step look ahead policy, the robot sees this landmark and dramatically changes the way in which it navigates back to the LRA. This helps the robot to achieve a much longer and more information rich path. Conversely, in the case of IC3 in Fig. (\ref{fig:LookSweep}a), the robot simply takes one more exploratory step along $\mathcal{G}$, but does not fundamentally change its tail strategy. Figure \eqref{fig:LookSweep} also implies that, for a high density of localization landmarks, such large differences in expected entropy reduction would not be seen between increments of the horizon $T_1$. This is intuitive because, with highly dense landmarks, regional values of $B_{\mathrm{pos}}$ would be similar, and the robot would not favor one area of the map over another due to localization.

\begin{figure}
\centering
\subfloat[][$T_1=1$]{\includegraphics[scale=.448]{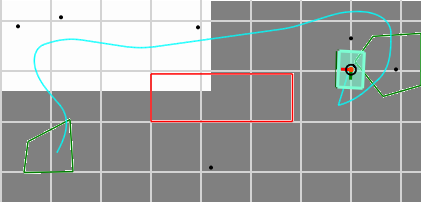}}

\centering
\subfloat[][$T_1=3$]{\includegraphics[scale=.448]{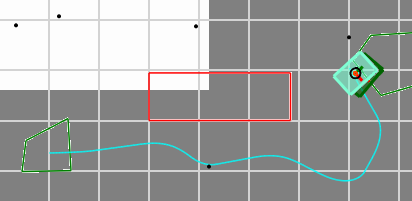}}

\caption{Two cases of the same obstacle scenario with the same initial conditions (IC3), but differing time horizons $T_1$. $\beta = 0.1$. }
\label{fig:LookExamplesOneObs}
\end{figure}

\subsection{Relationship between environmental complexity and look ahead distance}
Finally, consider the effect of obstacles on the RH-PIE and G-PIE. Figure \eqref{fig:LookExamplesOneObs} exemplifies this relationship.In Fig. \eqref{fig:LookExamplesOneObs}, by increasing the look ahead distance $T_1$, the robot is able to find a more information rich path in a different homotopy class; below the obstacle instead of above the obstacle.

The change in path occurs because shortest path computation of $R_\mathrm{tail}$ can begin from nodes which are within sensor range of the bottom central landmark. Even though the path in Fig. (\ref{fig:LookExamplesOneObs}a) is part of the subset of paths found for $T_1 =3$, the path in Fig. (\ref{fig:LookExamplesOneObs}b) is more information rich while still being feasible. This dramatic change, is similar to that seen by varying $\beta$, with the exception that the tail path remains localization seeking at $\beta = 0.1$. Thus, no looping behavior like that seen in Fig. (\ref{fig:BetaExamples}b) occurs. This is despite the fact that the loop in the bottom center shown in Fig. (\ref{fig:BetaExamples}b) exists in Fig. (\ref{fig:LookExamplesOneObs}b).

\subsubsection{Computation}
Finally, it is important to note the near-real-time performance of the RH-PIE algorithm. All code was written in C\# on a Windows 7 operating system. All code was run on a mobile I5 Intel Sandy Bridge processor. For $T_1 = 1$, the RH-PIE algorithm is able to calculate appropriate edge weights for $\mathcal{G}$ and plan a path in $\approx$ 30sec for a graph with 80 nodes. This is done while the robot is also processing sensor data, performing obstacle avoidance, and running visualization software. In contrast, the full G-PIE algorithm required $\approx$ 20min on the same hardware and a graph of 25 nodes, while not processing any auxiliary data. With a look ahead distance of 2, the RH-PIE algorithm is provides a solution in $\approx$ 1.5min while a distance of 3 requires $\approx$ 5min. This timing data implies that code optimization and a GPU implementation of the RH-PIE algorithm will allow real time replanning. 

\subsubsection{Summary and Discussion}
The RH-PIE algorithm performs in an intuitive manner, trading localization and exploration as a function of $\beta$, as shown in Fig. \eqref{fig:BetaExamples}. The optimality of the solution is traded in favor of practical speed as a function of the look ahead distance, exemplified by Fig. \eqref{fig:LookExamples}.

The tail reward function presented here provides a beneficial trade between conflicting objectives, constraining their values to guarantee tail optimality (in terms of shortest paths over $\mathcal{G}$), as in Fig. (\ref{fig:BetaExamples}a). By loosening this restriction, the RH-PIE algorithm is able to return longer paths which more fully explore the space, as shown in Fig. (\ref{fig:BetaSweep}, \ref{fig:BetaExamples}). In addition, careful implementation of shortest path algorithms can guarantee that the path, $X_{\mathrm{best}}$, is returned in finite (polynomial) time. Thus, even though the tail $X_{T_1:T}$ is suboptimal, loosening the constraints on $\beta $ provides informative paths while still guaranteeing the robot can re-localize upon path termination.

While locally optimal, the RH-PIE algorithm cannot make theoretical guarantees on global optimality due to the non-additive, path dependent nature of information. The RH-PIE still enables near real time performance with short look ahead distances. In addition, its simplicity and parallelizable structure allows for the majority of computation time (over 80\%) to be simple matrix manipulation. Thus, real time re-planning can currently be achieved by leveraging parallelism and GPU acceleration.

 The results in Fig. (\ref{fig:ExperimentalAchievement}) provide evidence that the assumption that successful path completion is dominated by terminal covariance, as claimed in \cite{Prentice}, is not well studied. Intermediate uncertainty in robot position, obstacle collision probability, and the interaction between obstacle avoidance and algorithmic guarantees are paramount in achieving practical conservatism in both the G-PIE and RH-PIE algorithms. The graph $\mathcal{G}$ in both algorithms could be extended to ensure obstacle avoidance using chance constraints similar to that presented in \cite{ChanceRRT}.

\section{CONCLUSIONS}
An information exploration planner, the Guaranteed Probabilistic Information Explorer (G-PIE) has been presented. The G-PIE algorithm solves the Integrated Exploration (IE) problem with probabilistic guarantees of path completion and asymptotically optimal exploration. An information based reward function is developed using entropy, which provides the flexibility to include a variety of exploration objectives. A formal bound to the information reward function is also developed for partially known environments. This bound enables fast computation of path rewards, reducing the computation time of the general problem by a factor of $10^3$. A novel connection is made between the Hamiltonian Path problem and general exploration tasks which restrict allowable paths to a graph. Thus, all non-exhaustive exploration planners, such as belief planners, on general graphs $\mathcal{G}$ cannot provide any guarantee on exploration performance. Simulation results show that the G-PIE behaves in an intuitive manner, exploring the unknown area while fulfilling the required terminal localization constraint conservatively. 

A computationally tractable, locally optimal approximation algorithm (RH-PIE) is also developed. The RH-PIE algorithm uses a receding horizon approach to give a locally optimal, information rich path. The RH-PIE algorithm guarantees that any returned path satisfies a constraint on re-localization. In addition, the RH-PIE algorithm provides a polynomial time approximation to the NP-hard longest path problem for robotic information gathering by utilizing a tail reward approximation which balances robot localization and information gathering. This balance is crucial in maintaining low pose uncertainty throughout a path and thus helps ensure the proper operation of low level controllers which rely heavily on accurate state estimates. 

Real world experiments demonstrate that the RH-PIE algorithm is able to generate paths which are both informative and ensure re-localization in controlled environments. The RH-PIE tuning parameter, $\beta$, is able to effectively trade between exploration and localization while keeping computation low. At the same time, experiments imply that the assumption that path completion is dominated by terminal covariance is incorrect. This finding reinforces the hypothesis that any objective function must balance information gain and localization to be of practical use.

\bibliography{LGPIE_ref}
\bibliographystyle{ieeetr}

%The following line comments out the appendix
%\iffalse
\appendices
	
  \section{Proof of Thm. \ref{thm:appxLongest}} \label{sect:poofOfThm}
  \prf{Consider the Hamiltonian Path problem on $\mathcal{G}(V,E)$. Consider the modified graph $\mathcal{G}'(V',E')$ which is identical to $\mathcal{G}$ with the exception of an arbitrary node $v_i \in V$, which is replaced with $v'_{\mathrm{out}}$, $v'_{\mathrm{in}}$, where $v'_{\mathrm{out}}$ has all outgoing edges of $v_i$ and only one incoming edge $e'_{\mathrm{between}} = \lbrace  v'_{\mathrm{in}},  v'_{\mathrm{out}}\rbrace$. Similarly, $v'_{\mathrm{in}}$ has only the incoming edges of $v_i$ and the outgoing edge $e'_{\mathrm{between}}$. The decision question is:``Does the longest path between $v'_{\mathrm{out}}$ and $v'_{\mathrm{in}}$ contain all verities in $V'$?". Clearly, $\mathcal{G}$ has a Hamiltonian path \textit{iff} the answer is ``YES". Now suppose there exists a polynomial time algorithm to approximate the maximization in Eq. \ref{eq:Optim}  and \ref{eq:C} over $\mathcal{G}'$ within an error $\mathrm{e}_r \in \R$. Then, the PTA algorithm can solve the Hamiltonian Path problem on  $\mathcal{G}$ \cite{Kaerger}. But since $\mathcal{G}$ was arbitrary, the existence of such an algorithm implies that $\mathbf{P = NP}$  }

  \begin{figure}
  	\centering
  	\includegraphics[scale=.5]{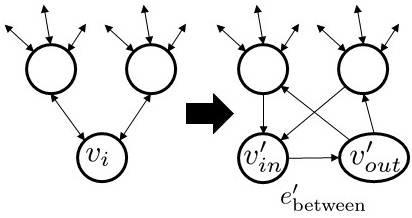}
  	\caption{Exemplification of the difference between $\mathcal{G}$ and $\mathcal{G}'$}
  \end{figure}

  \section{Derivation of Prop. \ref{prop:EntroBound}} \label{sect:proofOfEntroBound}
  Because of the cell independence assumption we just need an upper bound on $\E_Z[H(c|Z_1,...,Z_n)]$ as a function of the number of sample points $n$. To reduce notation, define $\tilde{Z} \equiv \{Z_1,...,Z_n\}$
  
  By careful manipulation of this expectation using the definition of entropy and Bayes rule, the following equivalence is evident:
  
\begin{equation} \label{prop:EntroBound1}
\E_Z[H(c|\tilde{Z})] =H(c) + \sum_{c \in \lb 0,1 \rb} \left[ p(c) H(\tilde{Z}|c) \right] - H(\tilde{Z})
\end{equation}
  
  Therefore, in order to upper bound this quantity, a lower bound on $H(\tilde{Z})$ is sufficient. For large $n$, the distribution on $(\tilde{Z}|c)$ is approximately normal, thus by the Moivre Laplace theorem:

\begin{equation} \label{prop:EntroBound2}
 H(\tilde{Z}|c) \approx {1 \over 2} \log \left( 2*\pi * e*n*\theta(1-\theta) \right) + \mathcal{O} (1/n)
\end{equation}

The assumption is made that the entropy of a Gaussian mixture approximating a mixture of binomials converges to the entropy of a mixture of binomials. Thus, the third term can be approximated using the entropy bound in \cite{EntroApxGM} as:

\begin{multline} \label{eq:EntroZBound}
H(\tilde{Z}) \geq \\ 
- \sum_{c} p(c) \log \left( \sum_{c '} p(c ') \mathcal{N} \bigg( \mu ',\mu, 2 n (\theta(1- \theta) \bigg) \right)
\end{multline}

\noindent
where $\mathcal{N}(x,y,\sigma)$ denotes a normal distribution evaluated at $y$ with mean $x$ and standard deviation $\sigma$. Here, $\mu = n \theta$ if $c=1$ and $\mu = n (1-\theta)$ otherwise. The same is true for $\mu '$ in terms of $c '$.  Denote $\Delta \mu = \mu_1 - \mu_0$, where $\mu_0 = n (1-\theta)  $ and $\mu_1 = n \theta$. Notice that the first term inside the log is simply the peak of a Normal distribution, thus the argument of the logarithm is:

\begin{multline}
\sum_{c ' \in \lb 0,1 \rb} p(c ')  \mathcal{N} \bigg( \mu,\mu ', 2 n (\theta(1-\theta) \bigg) \\ = {p(c) \over \sqrt{4 \pi n (\theta(1-\theta)}} + (1-\theta) \mathcal{N}(\Delta \mu , 2 n (\theta(1-\theta))
\end{multline}

\noindent
and

\begin{multline}
\mathcal{N}(\Delta \mu , 2 n (\theta(1-\theta)) \\ = {1 \over \sqrt{4 \pi n (\theta(1-\theta)}} exp(-{(2*n \theta -n)^2 \over n \theta (1-\theta)})$$ 
\end{multline}

By using the properties of logarithms and analyzing the asymptotic properties of the member functions, Eq. \eqref{eq:EntroZBound} implies that for large enough $n$:

\begin{equation} \label{prop:EntroBound3}
H(\tilde{Z}) \geq H(c) - \log \left({1 \over \sqrt{K*n}}\right)
\end{equation}
\noindent
where $K = 4\pi \theta (1-\theta)$ is a constant w.r.t $n$.

Thus, by combining Eqns. \eqref{prop:EntroBound1}, \eqref{prop:EntroBound2}, \eqref{prop:EntroBound3} and passing to the limit, the result follows:

\begin{multline}
\lim_{n \to \infty} H(c) + \sum_{c \in \lb 0,1 \rb} \left[ p(c) H(Z|c) \right] - H(Z) \geq \\ \lim_{n \to \infty}  \bigg[  H(c) + {1 \over 2} \log \Big( 2*\pi * e*n*\theta(1-\theta) \Big) + \mathcal{O} (1/n) \\  - H(c) + {1 \over 2} \log \left({1 \over K n}\right) \bigg]= {1 \over 2} \log({e \over 2}) 
\end{multline}

Thus the proof is complete if it can be shown that the entropy of a mixture of binomials approaches the entropy of a mixture of Gaussians. The of this is left out for brevity, but depends on discretion of the $\theta$ domain and careful application of the polynomial approximation theorem. \QED
  
\section{Proof of Prop. \ref{prop:EigenBound} } \label{sect:proofOfEigenBound}
In \cite{Prentice}, the authors show that in an EKF setting, the aggregate update equations along a robot's nominal path are:
$$\Sigma_{k+1} = L + G(\Sigma^{-1}_k +H^T Q^{-1}H)^{-1}G^T $$
\noindent
which resemble the standard Kalman filter equations. In the case where only one propagation/measurement pair is taken along an edge $e_k$, this equation reduces to the standard Kalman equations. 

In the \cite{knutson2001honeycombs}, Allen Knutson and Terrance Tao describe properties of a sum of Hermitian matrices. Suppose $A+B=C$. One property states that if $\lambda_i$, $\mu_i$ and $\nu_i$ are the $i$th eigenvalues of $A$, $B$ and $C$ respectively then:

$$\nu_{i+j+1} \leq \lambda_{i+1} + \mu_{j+1}$$

More specifically: $\nu_1 \leq \lambda_1 + \mu_1$. 

Because $Q$ and $\Sigma_k$ are assumed to be positive definite Hermitian:

$$(\Sigma_k^{-1}+ H^T Q^{-1} H)^{-1} \preccurlyeq \Sigma_k, (H^T Q^{-1} H)^{-1}$$

By these two observations, Eq. \eqref{eq:BRMKalman} is bounded by:

\begin{multline}
\lambda_1(\Sigma_{k+1}) \leq \lambda_1(L) + \\ \min\{\lambda_1(G(\Sigma^{-1}_k)G^T), \lambda_1(G(H^T Q^{-1}H)^{-1})G^T\}
\end{multline}

where $\lambda_1(\centerdot)$ is the largest eigenvalue of its argument. \QED

It is important to note that this bound becomes tight as $Q$ becomes small. This makes intuitive sense because $Q$ small implies a near perfect observation. Notice that this bound relies on the invertability of $(H^T Q^{-1}H)$, or equivalently the full rank of $H$. This is an assumption on the observability of the robotic system along edge $e_k$. In general, this is not true in sparse landmark environments. The proof of Lemma 4.2 can be done by inspection.

%\fi

\addtolength{\textheight}{-12cm}   % This command serves to balance the column lengths

%%%%%%%%%%%%%%%%%%%%%%%%%%%%%%%%%%%%%%%%%%%%%%%%%%%%%%%%%%%%%%%%%%%%%%%%%%%%%%%%

%%%%%%%%%%%%%%%%%%%%%%%%%%%%%%%%%%%%%%%%%%%%%%%%%%%%%%%%%%%%%%%%%%%%%%%%%%%%%%%%

%%%%%%%%%%%%%%%%%%%%%%%%%%%%%%%%%%%%%%%%%%%%%%%%%%%%%%%%%%%%%%%%%%%%%%%%%%%%%%%%

%%%%%%%%%%%%%%%%%%%%%%%%%%%%%%%%%%%%%%%%%%%%%%%%%%%%%%%%%%%%%%%%%%%%%%%%%%%%%%%%

\end{document}